\pdfoutput=1
\documentclass{article}
\usepackage{arxiv}
\usepackage{times}
\usepackage{natbib}

\usepackage{amsmath,amsfonts,bm}

\def\eqref#1{equation~\ref{#1}}
\def\1{\bm{1}}

\DeclareMathAlphabet{\mathsfit}{\encodingdefault}{\sfdefault}{m}{sl}
\SetMathAlphabet{\mathsfit}{bold}{\encodingdefault}{\sfdefault}{bx}{n}

\usepackage{hyperref}
\usepackage{url}
\usepackage{booktabs}     % tables
\usepackage{graphicx}     % figures
\usepackage{placeins}     % \FloatBarrier
\usepackage{subcaption}
\usepackage{multirow}
\usepackage{amsmath,amssymb}
\usepackage{xcolor}
\usepackage{tabularray}   % Table 2: row tints and cell shading without colortbl
\UseTblrLibrary{booktabs}
\usepackage{enumitem}
\usepackage{algorithm}
\usepackage{algpseudocode}
\usepackage{tikz}
\usetikzlibrary{arrows.meta,positioning,fit,backgrounds,calc,shapes.geometric,shapes.arrows}

\definecolor{cBlue}{HTML}{2A78D6}
\definecolor{cOrange}{HTML}{EB6834}
\definecolor{cAqua}{HTML}{1BAF7A}
\definecolor{cInk}{HTML}{0B0B0B}
\definecolor{cMuted}{HTML}{52514E}
\definecolor{cFaint}{HTML}{F4F4F1}

\makeatletter
\g@addto@macro\UrlBreaks{\UrlOrds}
\makeatother

\definecolor{cTie}{HTML}{8F8E8A}
\DeclareRobustCommand{\markfilled}{\tikz[baseline=-0.55ex]\fill[cMuted] (0,0) circle (0.45ex);}
\DeclareRobustCommand{\markhollow}{\tikz[baseline=-0.55ex]\draw[cMuted, line width=0.5pt] (0,0) circle (0.42ex);}
\DeclareRobustCommand{\marktie}{\tikz[baseline=-0.55ex]\draw[cTie, line width=0.8pt] (0,0) -- (1.1em,0);}

\title{\Large Rethinking Tabular Foundation Models On Data Streams}
\renewcommand{\shorttitle}{Rethinking Tabular Foundation Models On Data Streams}

\author{%
  Nilesh Verma$^{1}$ \quad Daniel Nowak-Assis$^{2}$ \quad Afonso Louren\c{c}o$^{3}$ \quad Albert Bifet$^{1}$ \\
  \textbf{Bernhard Pfahringer$^{1}$ \quad Maroua Bahri$^{2}$ \quad Nick Jin Sean Lim$^{1}$} \\[0.8em]
  \normalfont\small $^{1}$AI Institute, University of Waikato, New Zealand \\
  \normalfont\small $^{2}$LIP6, CNRS, Sorbonne Universit\'{e}, France \\
  \normalfont\small $^{3}$GECAD, Polytechnic of Porto, Portugal
}
\date{}

\begin{document}

\maketitle

\begin{abstract}
Tabular foundation models (TFMs) outperform established machine learning models on tabular benchmarks through in-context learning. Building on this success, interest is growing in applying them to data streams, where data arrive continuously and evolve over time. On a stream, a TFM adapts by updating its context rather than its parameters, so its accuracy and cost depend on which examples it keeps and how often it rebuilds its context. We therefore present a systematic study of TFMs on data streams, covering memory management, computational cost, and stream-specific challenges such as concept drift and delayed labels. We find that TFMs achieve the highest predictive performance, and that simply retaining the most recent examples is as effective as existing memory management techniques. They also recover faster than streaming learners after drift and keep the highest accuracy under label delay. This accuracy, however, comes at a high serving cost, since a nearly unchanged context is re-encoded at every prediction. These results point to architectural efficiency as the way forward for in-context stream learning.
\end{abstract}

\section{Introduction}

Data streams arise across a wide range of tabular applications, from fraud and intrusion detection to sensor monitoring and clinical decision support~\citep{dalpozzolo2014fraud,shyaa2024ids,arora2024review}. Unlike batch learning, stream learning must predict and adapt continuously under bounded memory and time, while the data-generating distribution changes~\citep{gama2014survey,bifet2018book}. Numerous stream learners have been proposed, from incrementally grown trees and drift-triggered ensembles~\citep{domingos2000vfdt,gomes2017arf,gomes2019srp,bifet2010leveraging} to online AutoML systems that adapt the whole pipeline~\citep{celik2023oaml,kulbach2022evoautoml,verma2024asml}, each adapting by updating model parameters as data arrives. Tabular foundation models (TFMs) offer a different route, predicting from labelled examples held as context in a single forward pass without updating parameters~\citep{muller2022pfn,hollmann2025tabpfnv2}, so adaptation to drift is handled entirely by updating the context memory. Recent approaches such as DualFIFO~\citep{lourenco2025ltm} and CURE~\citep{lee2026bounded} exploit this property and report large gains over streaming ensembles under a prequential protocol~\citep{gama2013prequential}. However, their evaluations are narrow in the streams and costs they cover, leaving no controlled account of how backbone, memory, label delay, and drift jointly determine predictive performance and serving cost.

\begin{figure}[t]
    \centering
    \resizebox{0.98\textwidth}{!}{% Overview, pictorial version. Three soft rounded panels, each dominated by a drawing
% rather than text: the stream as a row of instance cards with a drift break, each
% paradigm as a picture of what it updates (tree parameters, a re-selected pipeline,
% a FIFO memory in front of a frozen model), and the protocol as a predict/observe
% timeline feeding two result cards. Formulas sit as captions under the pictures.
% Needs colour macros cBlue, cAqua, cOrange, cInk, cMuted, cFaint and the tikz
% libraries positioning, fit, arrows.meta, calc, backgrounds, shapes.geometric.
\begin{tikzpicture}[
  font=\small,
  >={Stealth[length=4pt,width=3.5pt]},
  panel/.style={draw=cMuted!55, line width=.9pt, rounded corners=10pt, inner sep=0pt,
                fill=cFaint!60},
  sub/.style={draw=cMuted!45, line width=.7pt, rounded corners=7pt, inner sep=0pt, fill=white},
  ptitle/.style={font=\large\bfseries, inner sep=0pt},
  stitle/.style={font=\small\bfseries, inner sep=0pt},
  lbl/.style={font=\footnotesize, color=cMuted, align=center, inner sep=0pt},
  eq/.style={font=\small, align=center, inner sep=0pt},
  num/.style={circle, fill=#1, text=white, inner sep=0.5pt, minimum size=12pt,
              font=\footnotesize\bfseries},
  node/.style={circle, draw=#1!80!black, fill=#1!30, line width=.6pt, minimum size=9pt, inner sep=0pt},
  cell/.style={draw=#1!80!black, fill=#1!22, line width=.5pt},
  pipe/.style={draw=#1!80!black, fill=#1!22, line width=.6pt, rounded corners=2pt,
               minimum width=8mm, minimum height=4mm, inner sep=0pt, font=\tiny},
  arr/.style={->, draw=cInk, line width=.7pt},
  bigarr/.style={single arrow, draw=none, fill=cOrange!30, single arrow head extend=2.5pt,
                 minimum height=4.2mm, minimum width=5.5mm, inner sep=0pt},
  card/.style={draw=#1!75!black, fill=#1!15, line width=.8pt, rounded corners=4pt,
               align=center, inner sep=4pt, text width=2.0cm},
]
\def\ytop{2.9}\def\ybot{-2.2}
\begin{scope}[on background layer]
  \node[panel, fit={(0,\ytop) (3.7,\ybot)}] (P1) {};
\end{scope}
\begin{scope}[on background layer]
  \node[panel, fit={(4.15,\ytop) (17.1,\ybot)}] (P2) {};
\end{scope}
\begin{scope}[on background layer]
  \node[panel, fit={(17.55,\ytop) (22.85,\ybot)}] (P3) {};
\end{scope}
\begin{scope}[on background layer]
  \node[sub, fit={(0.15,2.1) (3.55,-1.9)}] (C0) {};
\end{scope}
\begin{scope}[on background layer]
  \node[sub, fit={(4.3,2.1) (8.4,-1.9)}] (C1) {};
\end{scope}
\begin{scope}[on background layer]
  \node[sub, fit={(8.6,2.1) (12.6,-1.9)}] (C2) {};
\end{scope}
\begin{scope}[on background layer]
  \node[sub, fit={(12.8,2.1) (16.9,-1.9)}] (C3) {};
\end{scope}
\begin{scope}[on background layer]
  \node[sub, fit={(17.7,2.1) (22.7,-1.9)}] (C4) {};
\end{scope}

% =============================================================================
% PANEL 1 : Data stream
% =============================================================================
\node[ptitle] at (1.85,\ytop-0.35) {Data stream};

% a queue of instance cards; colour switches at the drift point
\begin{scope}[shift={(0.35,0.15)}]
  \foreach \i/\c in {0/cBlue,1/cBlue,2/cBlue,3/cOrange,4/cOrange} {
    \begin{scope}[shift={(0.62*\i,0)}]
      \draw[cell=\c, rounded corners=1.2pt] (0,0) rectangle (0.50,1.10);
      \foreach \y in {0.26,0.55,0.84} \draw[draw=\c!70!black, line width=.45pt] (0.06,\y) -- (0.44,\y);
      \draw[draw=\c!70!black, line width=.45pt] (0.25,0.08) -- (0.25,1.02);
    \end{scope}
  }
  \draw[dashed, draw=cOrange!90!black, line width=.8pt] (1.80,-0.15) -- (1.80,1.25);
  \node[lbl, color=cOrange!90!black, anchor=south] at (1.80,1.28) {$p_t \neq p_{t+1}$};
  \draw[->, draw=cMuted, line width=.6pt] (-0.15,-0.25) -- (2.90,-0.25) node[lbl, below=2pt, xshift=-3pt] {$t$};
\end{scope}
\node[eq,font=\normalsize] at (1.85,-1.30) {$(\mathbf{x}_t,y_t)\sim p_t$};

% =============================================================================
% PANEL 2 : Three ways to adapt
% =============================================================================
\node[ptitle] at (10.6,\ytop-0.35) {Adaptation strategies};

% ---- column 1 : classical incremental -----------------------------------------
\node[stitle] at (6.35,1.75) {\tikz[baseline=-3pt]\node[num=cBlue]{1}; Classical streaming};
\begin{scope}[shift={(6.35,0.45)}]
  % an instance enters a model whose parameters are rewritten every step
  \draw[cell=cBlue, rounded corners=1.2pt] (-1.45,-0.30) rectangle (-0.95,0.30);
  \foreach \y in {-0.12,0.0,0.12} \draw[draw=cBlue!70!black, line width=.45pt] (-1.39,\y) -- (-1.01,\y);
  \node[lbl] at (-1.20,-0.55) {$(\mathbf{x}_t,y_t)$};
  \draw[->, draw=cInk, line width=.6pt] (-0.90,0.0) -- (-0.55,0.0);
  \draw[draw=cBlue!85!black, fill=cBlue!12, line width=.7pt, rounded corners=2pt]
    (-0.5,-0.42) rectangle (0.5,0.42);
  \node[font=\footnotesize] at (0,0.0) {$\theta_t$};
  % update loop on the parameters
  \draw[->, draw=cBlue!85!black, line width=.7pt] (0.85,0.2) arc (200:-110:0.22);
  \node[lbl] at (1.10,0.75) {$\theta_{t+1}$};
  \draw[->, draw=cMuted, line width=.5pt] (0,0.42) -- (0,0.75) node[lbl, above=-1pt] {$\hat y_t$};
\end{scope}
\node[eq] at (6.35,-1.3) {$\theta_{t+1} \leftarrow U\!\left(\theta_t,(\mathbf{x}_t,y_t)\right)$};

% ---- column 2 : streaming AutoML ----------------------------------------------
\node[stitle] at (10.6,1.75) {\tikz[baseline=-3pt]\node[num=cAqua!90!black]{2}; Streaming AutoML};
\begin{scope}[shift={(10.6,0.45)}]
  \foreach \j/\o/\c in {0/0.6/cMuted, 1/0/cAqua, 2/-0.6/cMuted} {
    \node[pipe=\c] (p\j a) at (-0.75,\o) {prep};
    \node[pipe=\c] (p\j b) at (0.35,\o) {model};
    \draw[->, draw=\c!80!black, line width=.5pt] (p\j a) -- (p\j b);
  }
  \draw[draw=cAqua!90!black, line width=1.1pt, rounded corners=3pt]
    (-1.3,-0.28) rectangle (0.9,0.28);
  \draw[->, draw=cAqua!90!black, line width=.7pt] (1.38,0.78) -- (1.02,0.08);
  \node[lbl, color=cAqua!60!black, anchor=east] at (1.72,0.97) {$\arg\max$};
  \end{scope}
\node[eq] at (10.6,-1.3) {$\Pi_{t+1} \leftarrow \arg\max\limits_{\Pi \in \mathcal{P}} \mathrm{score}(\Pi)$};

% ---- column 3 : in-context foundation models ----------------------------------
\node[stitle] at (14.85,1.75) {\tikz[baseline=-3pt]\node[num=cOrange!95!black]{3}; In-context models};
\begin{scope}[shift={(14.85,0.45)}]
  % FIFO memory
  \foreach \i in {0,...,4}
    \draw[cell=cOrange] (-1.45+0.22*\i,-0.12) rectangle ++(0.19,0.24);
  \draw[->, draw=cMuted, line width=.5pt] (-1.65,0.0) -- (-1.48,0.0);
  \node[lbl] at (-0.95,-0.4) {$M_t$};
  \draw[->, draw=cInk, line width=.6pt] (-0.3,0.0) -- (-0.05,0.0);
  % frozen model block
  \draw[draw=cOrange!85!black, fill=cOrange!12, line width=.7pt, rounded corners=2pt]
    (0,-0.42) rectangle (1.0,0.42);
  \node[font=\footnotesize] at (0.5,0.1) {$\theta$};
  % snowflake
  \begin{scope}[shift={(0.5,-0.2)}]
    \foreach \a in {0,60,120} \draw[draw=cAqua!80!black, line width=.5pt] (\a:0.11) -- (\a+180:0.11);
  \end{scope}
    \draw[->, draw=cMuted, line width=.5pt] (0.5,0.42) -- (0.5,0.75) node[lbl, above=-1pt] {$\mathbf{x}_t$};
\end{scope}
\node[eq] at (14.85,-1.3) {$\hat p\!\left(y \mid \mathbf{x}_t, M_t\right),\quad \theta\ \text{fixed}$};

% =============================================================================
% PANEL 3 : One protocol, two axes
% =============================================================================
\node[ptitle] at (20.2,\ytop-0.35) {One protocol};

% predict then observe, on a timeline
\begin{scope}[shift={(20.2,1.42)}]
  \draw[->, draw=cMuted, line width=.6pt] (-2.1,-0.55) -- (2.1,-0.55) node[lbl, below=2pt, xshift=-3pt] {$t$};
  \node[draw=cBlue!75!black, fill=cBlue!15, rounded corners=6pt, font=\footnotesize, inner sep=3pt]
    (pr) at (-1.05,0) {predict $\hat y_t = f(\mathbf{x}_t)$};
  \node[draw=cOrange!80!black, fill=cOrange!15, rounded corners=6pt, font=\footnotesize, inner sep=3pt]
    (ob) at (1.15,0) {observe $y_t$};
  \draw[arr] (pr) -- (ob);
\end{scope}
\node[eq, font=\footnotesize] at (20.2,0.12) {$\lvert M_t \rvert \le B$,\ \ $q$ queries,\ \ delay $d$};

\node[card=cAqua] (acc) at (19.0,-1.42)
  {\textbf{Accuracy}};
\node[card=cOrange] (cost) at (21.4,-1.42)
  {\textbf{Cost}};
\draw[arr] (20.2,-0.22) -- (20.2,-0.62) -| (acc.north);
\draw[arr] (20.2,-0.62) -| (cost.north);

% =============================================================================
% flow between panels
% =============================================================================
\node[bigarr] at (3.925,0.35) {};
\node[bigarr] at (17.325,0.35) {};
\end{tikzpicture}}
    \caption{Approaches to adaptation in data streams, compared under a common prequential protocol measuring predictive performance and computational cost.}
    \label{fig:overview}
\end{figure}
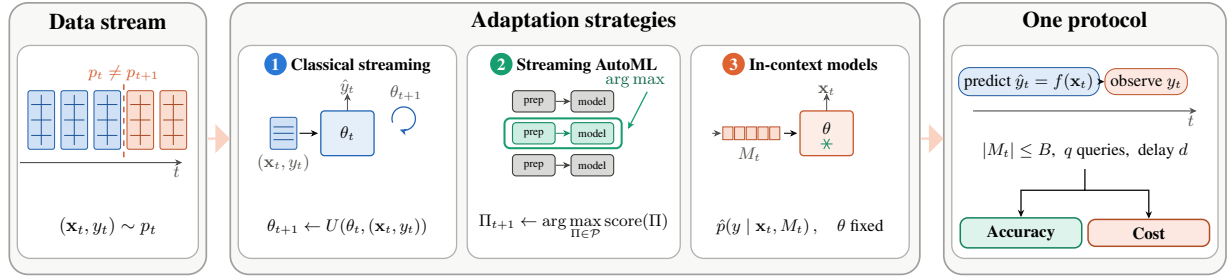

To close this gap, in this paper we systematically study TFMs on data streams. Our contributions are as follows:
\begin{itemize}[leftmargin=*,itemsep=1pt,topsep=2pt]
\item We compare TFM backbones against streaming learners and streaming AutoML systems under the same protocol on real and synthetic streams with different types of drift, and find that TFMs are the most accurate but also the most costly learners (Figure~\ref{fig:overview}).
\item We conduct a stream-specific analysis of memory composition, memory size, label delay, context refresh rate, and query batching, and find that backbone quality and memory size dominate predictive performance, while the memory management policy matters mainly under delayed labels.
\item We identify repeated context encoding as the dominant serving cost of streaming TFMs, and show with compact causal models that reusing encoded context keeps cost nearly constant across memory sizes while staying close to re-encoding in accuracy.
\end{itemize}

\section{Background}
\label{sec:background}

\textbf{Data streams.} A data stream is an unbounded sequence of examples $(\mathbf{x}_t,y_t)$ arriving one at a time, where $\mathbf{x}_t\in\mathcal{X}$ is a feature vector and $y_t\in\mathcal{Y}$ its class label. Each example is drawn from a distribution $p_t$ that may change over time, and such a change is a \emph{concept drift}, which may be abrupt, gradual, incremental, or recurring~\citep{gama2014survey,lu2019drift}. A stream classifier must use bounded memory and time per example, be ready to predict at any time, and adapt when $p_t$ changes~\citep{bifet2018book}. It is evaluated \emph{prequentially} (test-then-train), in which each example first tests the current model and only then becomes available for learning~\citep{gama2013prequential}, and in practice its label may arrive only after a delay of $d$ examples~\citep{delayed_labelling}. Most stream learners meet these requirements by updating the model as data arrives. Hoeffding trees grow incrementally from sufficient statistics~\citep{domingos2000vfdt,bifet2009hat}, drift detectors such as ADWIN trigger a response when the distribution shifts~\citep{bifet2007adwin}, and ensembles such as Adaptive Random Forest, Streaming Random Patches, and Leveraging Bagging replace degraded members~\citep{gomes2017arf,gomes2019srp,bifet2010leveraging}. SAM-kNN instead retains past examples in short- and long-term memories~\citep{losing2016samknn}, and is the classical learner closest to context-based adaptation. Online AutoML extends adaptation to the whole pipeline, re-running a search on detected drift~\citep{celik2023oaml}, evolving a population of pipelines~\citep{kulbach2022evoautoml}, or searching around the current best configuration~\citep{verma2024asml}.

\textbf{Tabular foundation models.} A tabular foundation model (TFM) is a transformer pretrained once on a large number of synthetic tables~\citep{hollmann2023tabpfn,google2026tabfm}. At prediction time, a context $D_{\mathrm{ctx}}=\{(\mathbf{x}_i,y_i)\}_{i=1}^{n}$ of labelled rows and a query $\mathbf{x}^{*}$ pass through a single forward pass, $\hat{p}(y\mid\mathbf{x}^{*},D_{\mathrm{ctx}})=f_{\theta}(\mathbf{x}^{*},D_{\mathrm{ctx}})$, and on small and medium tables TFMs now match or exceed tuned gradient-boosted trees~\citep{erickson2025tabarena}. TabPFN lets every context row attend to every other row directly~\citep{hollmann2025tabpfnv2}, while TabICL first embeds each row from its columns and then lets rows attend to one another~\citep{qu2026tabiclv2}, and TabFM alternates attention over columns and rows before attending over compressed row vectors~\citep{google2026tabfm}, so in all cases encoding the context costs time quadratic in $n$. Drift-Resilient TabPFN~\citep{helli2024drtabpfn} is the one backbone pretrained for shift, on temporally shifting priors with a known time covariate. Context construction has been studied for fixed tables, either by retrieving rows per query~\citep{thomas2024retrieval} or by compressing and pruning the context~\citep{feuer2024tunetables}, but all assume a fixed pool of rows drawn from a single distribution.

\textbf{Streaming TFMs.} In a stream, the context becomes a memory $M_t$ of at most $B$ labelled rows that must be \emph{maintained} as rows keep arriving, trading recent rows that allow fast adaptation against older rows that preserve earlier concepts~\citep{lourenco2025ltm}. The closest prior works (Appendix~\ref{app:prior}) are DualFIFO~\citep{lourenco2025ltm}, which splits the memory into a short-term FIFO buffer and a class-balanced long-term store, and CURE~\citep{lee2026bounded}, which evicts redundant rows to keep the memory informative. Their evaluations leave several gaps open. First, neither compares performance across drift types, since DualFIFO is evaluated only on real-world streams, where drift type is unknown, and CURE includes a single synthetic stream, with abrupt drift. Second, neither places serving cost beside that of classical learners. Because $M_t$ and $M_{t+1}$ differ by a single row, every step re-encodes almost the same set of rows, yet DualFIFO times ten instances per forward pass and CURE compares its per-step runtime only with DualFIFO. Third, both assume that each label arrives immediately, although labels in deployed streams often arrive late. These gaps motivate the design of our study.

\section{Experimental setup}
\label{sec:setup}

\textbf{Streams.} The evaluation covers real and synthetic classification streams (Table~\ref{tab:streams}). Several real streams are distributed by the USP data stream repository \citep{souza2020challenges}. Only datasets whose row order reflects the order in which observations arrived are included, since an arbitrary file order does not represent a true stream. The synthetic streams use data generators with abrupt, gradual, incremental, and recurring drift, the shapes drift takes over time \citep{lu2019drift}. As each drift type is paired with a different generator, they serve only to describe the stream conditions and not to compare them directly.

\begin{table}[t]
\centering
\caption{Evaluation streams.}
\label{tab:streams}
\small
\setlength{\tabcolsep}{4pt}
\begin{tabular}{llrrrl}
\toprule
Stream & Type & Instances & Features & Classes & Drift \\
\midrule
\texttt{electricity}~\citep{harries1999splice}  & real & 45{,}312  & 8  & 2  & unknown \\
\texttt{noaa}~\citep{elwell2011nse}             & real & 18{,}159  & 8  & 2  & unknown \\
\texttt{meter}~\citep{souza2020ladpu}           & real & 22{,}948  & 96 & 10 & unknown \\
\texttt{rialto}~\citep{losing2016samknn}        & real & 82{,}250  & 27 & 10 & unknown \\
\texttt{posture\_no8}~\citep{kaluza2010posture} & real & 163{,}477 & 3  & 10 & unknown \\
\midrule
\texttt{agr\_a}~\citep{agrawal1993database} & synth. & 30{,}000  & 9  & 2 & abrupt \\
\texttt{sea\_g}~\citep{street2001sea}       & synth. & 100{,}000 & 3  & 2 & gradual \\
\texttt{rbf\_m}~\citep{bifet2010moa}        & synth. & 100{,}000 & 10 & 5 & incremental \\
\texttt{tree\_r}~\citep{domingos2000vfdt}   & synth. & 60{,}000  & 10 & 2 & recurring \\
\bottomrule
\end{tabular}
\end{table}

\textbf{Methods.}
The comparison covers dedicated streaming learners, streaming AutoML, and in-context TFMs. Classical streaming learners include Hoeffding Adaptive Tree \citep{bifet2009hat}, Adaptive Random Forest \citep{gomes2017arf}, Streaming Random Patches \citep{gomes2019srp}, Leveraging Bagging \citep{bifet2010leveraging}, and SAM-kNN \citep{losing2016samknn}. Streaming AutoML methods include ASML \citep{verma2024asml}, EvoAutoML \citep{kulbach2022evoautoml}, and OAML \citep{celik2023oaml}. In-context TFMs include CURE \citep{lee2026bounded}, DualFIFO \citep{lourenco2025ltm}, and FIFO-based variants of TabICL~v2 \citep{qu2026tabiclv2}, TabPFN~v1 \citep{hollmann2023tabpfn}, Drift-Resilient TabPFN \citep{helli2024drtabpfn}, and TabFM \citep{google2026tabfm}, where TabPFN~v1 is also the backbone of DualFIFO and Drift-Resilient TabPFN. These TFMs span different backbone designs, pretraining strategies, and model scales, from models with tens of millions of parameters to a billion-scale backbone, allowing us to examine architecture, scale, and drift-oriented pretraining. No-Change and Majority-Class serve as simple reference baselines. Following established evaluation practice \citep{erickson2025tabarena,lourenco2025ltm}, all methods use the settings and implementations described in their original work, with streaming learners and streaming AutoML methods running on CPU and TFMs on GPU.

\textbf{Evaluation protocol and metrics.}
All methods follow the same prequential protocol, where each arriving instance is first predicted and becomes available for learning only after a label delay of $d$ instances \citep{gama2013prequential,delayed_labelling}. Memory-based methods retain at most $B$ labelled examples, and unless varied explicitly, all runs have no label delay $d$ and share the same budget $B$ across memory-based methods. We report accuracy together with balanced accuracy, which weights classes equally, and the chance-corrected statistics Cohen's $\kappa$, which adjusts for agreement expected by chance \citep{cohen1960}, and temporal $\kappa_t$, which adjusts for the accuracy of the No-Change baseline \citep{bifet2015kappat}. Latency is reported in milliseconds per instance, measured as wall-clock serving time on each method's own host, and memory growth per instance.

To support reproducibility, we release the full experimental pipeline used in this study, including the code, run configurations for every experiment, and the raw result files behind all reported tables and figures.\footnote{\url{https://github.com/TechyNilesh/StreamTFM-Study}}

\section{Results}
\label{sec:results}
\label{sec:experiments}

\textbf{Predictive performance and cost.} TabFM with FIFO ranks first in accuracy on all streams but one, ahead of both existing memory management systems and the streaming AutoML learners, with Drift-Resilient TabPFN narrowly ahead on the abrupt-drift stream (Table~\ref{tab:metrics}, with per-stream accuracies and a Wilcoxon--Holm comparison in Appendix~\ref{app:results}). The measured parameter counts separate scale from design. TabICL~v2 clearly exceeds TabPFN~v1 at almost the same parameter count, whereas TabFM adds only a small further gain for many times more parameters (Appendix~\ref{app:repro}). Backbone quality therefore reflects architecture and pretraining at least as much as scale. Drift-Resilient TabPFN makes the pretraining effect visible on its own, since at the size and architecture of TabPFN~v1 it gains several points on average, with large gains on meter, rialto and the recurring-drift stream and small changes elsewhere. Chance-corrected metrics mainly reorder the middle of the field and reveal where accuracy benefits from class imbalance or label autocorrelation. Electricity is the clearest case. Label autocorrelation lets No-Change reach a high accuracy there and some streaming learners fall below it, whereas TabFM+FIFO and TabICL+FIFO climb well above it and their temporal $\kappa_t$ is roughly twice that of the best ensemble, which is why we also report $\kappa_t$.

\begin{figure}[!ht]
    \centering
    \includegraphics[width=\linewidth]{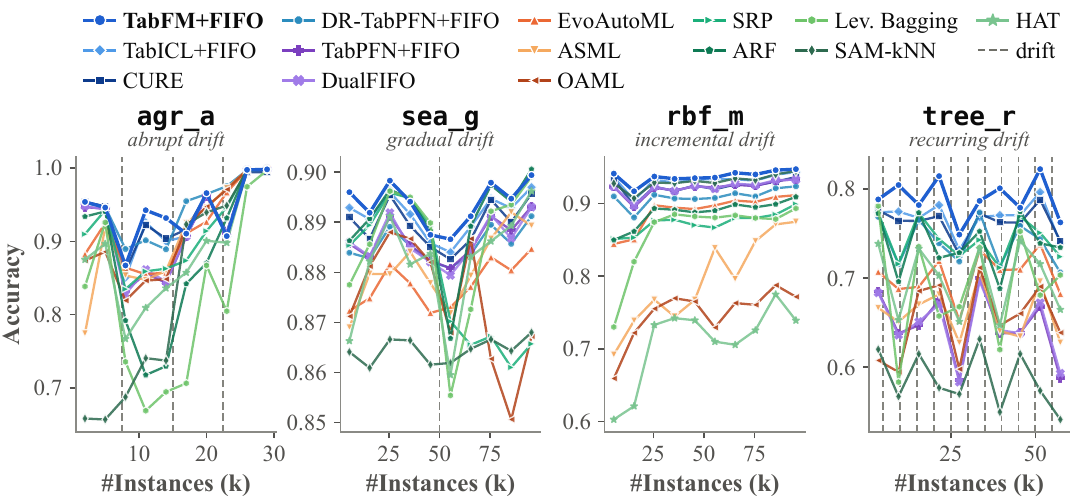}
    \caption{Accuracy over the synthetic streams, each with its own drift type, in stages of roughly a tenth of the stream. Dashes mark concept changes.}
    \label{fig:windows}
\end{figure}

Performance also varies across streams (Appendix~\ref{app:results-variation}), with SRP showing the smallest spread among non-trivial methods, while SAM-kNN and EvoAutoML are more dependent on the stream. Figure~\ref{fig:windows} shows that the TFM advantage appears early and persists across several concept changes, although it is not uniform over time. On the abrupt-drift stream, the in-context methods start at higher predictive performance and recover to higher levels after the first drift. In a later, strongly autocorrelated segment, SRP and EvoAutoML overtake TabFM+FIFO. The size of the advantage depends on the drift type, since gradual and incremental streams separate almost nobody, while under recurring drift the swing within a single method exceeds the gap between paradigms. These differences show why aggregate performance should be read alongside temporal behaviour.

\begin{figure}[!ht]
    \centering
    \includegraphics[width=\linewidth]{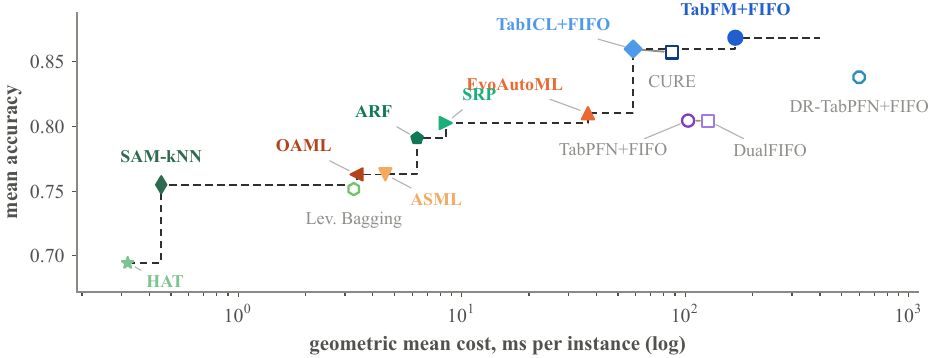}
    \caption{Mean accuracy against mean cost over all streams, coloured by method family. Filled marks (\markfilled) are on the Pareto front, hollow marks (\markhollow) are dominated, and grey ties (\marktie) join each memory management system to FIFO on the same backbone.}
    \label{fig:pareto}
\end{figure}

The predictive advantage comes at substantial computational cost (Figure~\ref{fig:pareto}). In-context methods occupy the most accurate and expensive end of the comparison, with the additional predictive gains requiring more than an order of magnitude higher latency than the cheapest ensemble, even though the TFMs run on GPUs. Because the asymmetry favours the TFMs, normalising for hardware would widen rather than close this gap, and the same ordering holds when times are converted to cloud cost (Appendix~\ref{app:cost-price}). Both existing memory management systems are slightly less accurate and more expensive than FIFO on their own backbone, motivating the controlled memory studies below. Latency also varies severalfold across streams within a method and is highest on the widest stream, so mean cost alone does not capture the full deployment demands.

\begin{table}[t]
\centering
\caption{Predictive performance and cost of all methods, averaged over streams. Rank is the per-stream accuracy rank, time uses the geometric mean, and memory is growth per instance. Best \textbf{in bold}, second \underline{underlined}.}
\label{tab:metrics}
\small
\definecolor{tintincontext}{HTML}{E3EDFA}
\definecolor{tintAutoML}{HTML}{FCEBE5}
\definecolor{tintclassical}{HTML}{E1F5EE}
\definecolor{tinttrivial}{HTML}{EAEAE9}
\definecolor{swtabfmfifo}{HTML}{1F5FD0}
\definecolor{swtabiclv2fifo}{HTML}{4F9AE8}
\definecolor{swCURE}{HTML}{0D3B8C}
\definecolor{swdrtabpfnfifo}{HTML}{2B8CBE}
\definecolor{swtabpfnv1fifo}{HTML}{7B3FBF}
\definecolor{swLTM}{HTML}{A07BE8}
\definecolor{swevoautoml}{HTML}{EB6834}
\definecolor{swsrp}{HTML}{1BAF7A}
\definecolor{swarf}{HTML}{0E7A54}
\definecolor{swleveragingbagging}{HTML}{6EC46A}
\definecolor{swasml}{HTML}{F5A95F}
\definecolor{swoaml}{HTML}{B2431A}
\definecolor{swsamknnb1000}{HTML}{2D6A4F}
\definecolor{swhoeffdingadaptivetree}{HTML}{79C48F}
\definecolor{swmajorityclass}{HTML}{9A9A95}
\definecolor{swnochange}{HTML}{5A5A55}
\definecolor{acc0}{HTML}{A6C6EE}
\definecolor{acc1}{HTML}{A7C7EE}
\definecolor{acc2}{HTML}{A8C8EE}
\definecolor{acc3}{HTML}{ACCAEF}
\definecolor{acc4}{HTML}{B2CEF0}
\definecolor{acc5}{HTML}{B2CEF0}
\definecolor{acc6}{HTML}{B1CEF0}
\definecolor{acc7}{HTML}{B2CFF0}
\definecolor{acc8}{HTML}{B5D0F1}
\definecolor{acc9}{HTML}{BDD5F2}
\definecolor{acc10}{HTML}{BAD3F2}
\definecolor{acc11}{HTML}{BAD3F2}
\definecolor{acc12}{HTML}{BCD4F2}
\definecolor{acc13}{HTML}{C8DCF4}
\definecolor{acc14}{HTML}{FCFDFE}
\definecolor{acc15}{HTML}{FFFFFF}
\resizebox{\linewidth}{!}{%
\begin{tblr}{colspec={lrrrrrrrrrrrrrrr}, colsep=3pt, rowsep=0.6pt, cell{2}{2-16}={halign=c},
  row{3}={bg=tintincontext}, cell{3}{4}={bg=acc0},
  row{4}={bg=tintincontext}, cell{4}{4}={bg=acc1},
  row{5}={bg=tintincontext}, cell{5}{4}={bg=acc2},
  row{6}={bg=tintincontext}, cell{6}{4}={bg=acc3},
  row{7}={bg=tintincontext}, cell{7}{4}={bg=acc4},
  row{8}={bg=tintincontext}, cell{8}{4}={bg=acc5},
  row{9}={bg=tintAutoML}, cell{9}{4}={bg=acc6},
  row{10}={bg=tintclassical}, cell{10}{4}={bg=acc7},
  row{11}={bg=tintclassical}, cell{11}{4}={bg=acc8},
  row{12}={bg=tintclassical}, cell{12}{4}={bg=acc9},
  row{13}={bg=tintAutoML}, cell{13}{4}={bg=acc10},
  row{14}={bg=tintAutoML}, cell{14}{4}={bg=acc11},
  row{15}={bg=tintclassical}, cell{15}{4}={bg=acc12},
  row{16}={bg=tintclassical}, cell{16}{4}={bg=acc13},
  row{17}={bg=tinttrivial}, cell{17}{4}={bg=acc14},
  row{18}={bg=tinttrivial}, cell{18}{4}={bg=acc15}}
\toprule
& \SetCell[c=2]{c} Rank & & \SetCell[c=3]{c} Accuracy & & & \SetCell[c=2]{c} Balanced acc. & & \SetCell[c=2]{c} Cohen's $\kappa$ & & \SetCell[c=2]{c} $\kappa_t$ & & \SetCell[c=2]{c} Time (ms/inst.) & & \SetCell[c=2]{c} Memory (KB/inst.) & \\
\cmidrule[lr]{2-3} \cmidrule[lr]{4-6} \cmidrule[lr]{7-8} \cmidrule[lr]{9-10} \cmidrule[lr]{11-12} \cmidrule[lr]{13-14} \cmidrule[lr]{15-16}
Method & mean & med. & mean & med. & std. & mean & rank & mean & rank & mean & rank & mean & med. & mean & med. \\
\midrule
{\color{swtabfmfifo}\rule{0.85ex}{0.85ex}}\,TabFM + FIFO & \textbf{1.11} & \textbf{1} & \textbf{.869} & \textbf{.924} & .104 & \textbf{.829} & \textbf{1.33} & \textbf{.772} & \textbf{1.22} & \textbf{.730} & \textbf{1.11} & 167 & 129 & 4.80 & 1.50 \\
{\color{swtabiclv2fifo}\rule{0.85ex}{0.85ex}}\,TabICL + FIFO & \underline{2.56} & \underline{2} & \underline{.860} & .906 & .105 & \underline{.820} & \underline{2.89} & \underline{.758} & \underline{2.67} & \underline{.712} & \underline{2.56} & 58.4 & 71.4 & 3.73 & 1.81 \\
{\color{swCURE}\rule{0.85ex}{0.85ex}}\,CURE & 2.89 & 3 & .857 & \underline{.907} & .106 & .819 & \underline{2.89} & .754 & 2.78 & .704 & 2.89 & 87.3 & 97.9 & 5.94 & 1.96 \\
{\color{swdrtabpfnfifo}\rule{0.85ex}{0.85ex}}\,DR-TabPFN + FIFO & 5.33 & 5 & .838 & .868 & \underline{.103} & .794 & 5.67 & .724 & 5.33 & .659 & 5.33 & 599 & 542 & 5.38 & 1.78 \\
{\color{swtabpfnv1fifo}\rule{0.85ex}{0.85ex}}\,TabPFN + FIFO & 6.89 & 6 & .805 & .814 & .122 & .762 & 7.33 & .677 & 7.44 & .619 & 6.89 & 103 & 89.8 & 2.13 & 1.32 \\
{\color{swLTM}\rule{0.85ex}{0.85ex}}\,DualFIFO & 6.89 & 6 & .804 & .814 & .122 & .770 & 5.67 & .680 & 6.22 & .619 & 6.89 & 127 & 116 & 5.63 & 2.21 \\
{\color{swevoautoml}\rule{0.85ex}{0.85ex}}\,EvoAutoML & 7.67 & 7 & .810 & .878 & .120 & .761 & 8.00 & .679 & 7.56 & .627 & 7.67 & 36.7 & 22.9 & 7.58 & 2.67 \\
{\color{swarf}\rule{0.85ex}{0.85ex}}\,ARF & 8.00 & 9 & .791 & .798 & .104 & .742 & 8.22 & .658 & 8.00 & .592 & 8.00 & 6.30 & 6.50 & 6.93 & 6.87 \\
{\color{swsrp}\rule{0.85ex}{0.85ex}}\,SRP & 8.11 & 8 & .803 & .802 & \textbf{.100} & .749 & 9.22 & .670 & 8.44 & .605 & 8.11 & 8.47 & 8.50 & 10.7 & 7.15 \\
{\color{swleveragingbagging}\rule{0.85ex}{0.85ex}}\,Lev.\ Bagging & 10.33 & 11 & .752 & .785 & .122 & .707 & 9.78 & .601 & 10.22 & .525 & 10.33 & 3.29 & 2.64 & 5.69 & 4.24 \\
{\color{swasml}\rule{0.85ex}{0.85ex}}\,ASML & 10.67 & 11 & .763 & .778 & .126 & .719 & 10.44 & .616 & 10.67 & .550 & 10.67 & 4.55 & 3.58 & \textbf{0.97} & \underline{0.27} \\
{\color{swoaml}\rule{0.85ex}{0.85ex}}\,OAML & 11.22 & 11 & .763 & .783 & .127 & .720 & 10.78 & .613 & 11.11 & .540 & 11.22 & 3.38 & 1.94 & 4.11 & 1.70 \\
{\color{swsamknnb1000}\rule{0.85ex}{0.85ex}}\,SAM-kNN & 11.22 & 13 & .755 & .817 & .134 & .708 & 10.78 & .582 & 11.22 & .472 & 11.22 & 0.45 & 0.38 & 4.86 & 2.93 \\
{\color{swhoeffdingadaptivetree}\rule{0.85ex}{0.85ex}}\,HAT & 12.33 & 14 & .694 & .709 & .166 & .654 & 12.22 & .530 & 12.33 & .430 & 12.33 & 0.32 & 0.31 & 6.34 & 3.79 \\
{\color{swmajorityclass}\rule{0.85ex}{0.85ex}}\,Majority-Class & 15.22 & 15 & .431 & .523 & .249 & .323 & 15.67 & -.011 & 15.67 & -.139 & 15.22 & \underline{0.19} & \underline{0.16} & 4.48 & \textbf{0.18} \\
{\color{swnochange}\rule{0.85ex}{0.85ex}}\,No-Change & 15.56 & 16 & .413 & .505 & .306 & .380 & 15.11 & .112 & 15.11 & .000 & 15.56 & \textbf{0.17} & \textbf{0.12} & \underline{1.41} & 0.56 \\
\bottomrule
\end{tblr}}

\end{table}

\begin{figure}[!ht]
    \centering
    \includegraphics[width=0.95\linewidth]{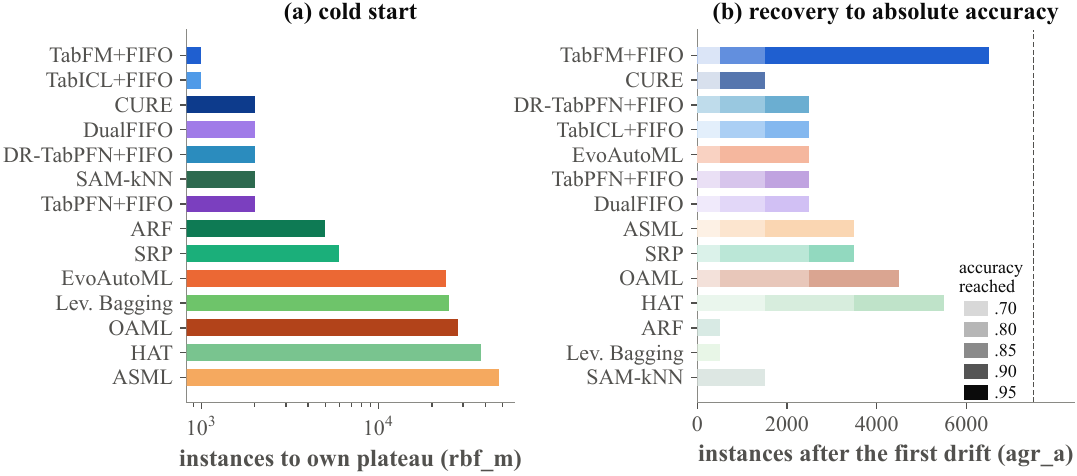}
    \caption{Instances to reach the initial plateau (left) and to regain accuracy levels after the first abrupt drift in \texttt{agr\_a} (right). A missing bar marks a level not reached before the next drift (dashed line).}
    \label{fig:dynamics}
\end{figure}

\begin{figure}[!ht]
    \centering
    \includegraphics[width=0.95\linewidth]{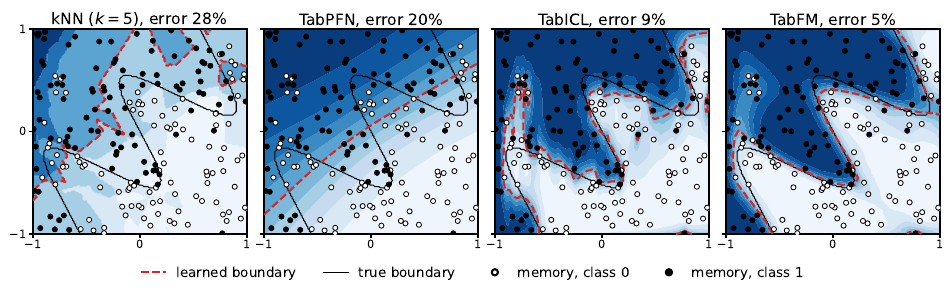}
    \caption{Decision functions of kNN and the TFMs on the same FIFO memory shortly after a gradual drift, with the misclassified share of the input area in each title.}
    \label{fig:boundary-main}
\end{figure}

\textbf{Cold start and drift recovery.} Figure~\ref{fig:dynamics} shows how quickly methods reach their initial plateau and fixed accuracy levels after the first abrupt drift. In-context methods begin near their plateau, while streaming ensembles and AutoML learners need longer, since a tree ensemble must see enough instances for its split tests, whereas an in-context model reads the concept directly from the rows in its memory. SAM-kNN, which also adapts through stored examples, is the exception among streaming learners. Figure~\ref{fig:boundary-main} makes this visible on a synthetic stream with irrelevant features, where kNN and the TFMs share the same FIFO memory. The irrelevant features corrupt the distances of kNN, while the TFMs ignore them and follow the true boundary in the order of their overall accuracy (Appendix~\ref{app:boundary}). Until the next concept change, every in-context method recovers quickly, with TabFM+FIFO recovering furthest, while the AutoML learners recover more slowly and level off lower. SRP and HAT recover only late, and ARF, Leveraging Bagging and SAM-kNN never regain even the lowest level shown. Later drifts on this stream lead to easier concepts and are less informative.

%\textbf{Controlled studies.}
The comparison above does not isolate the contributions of backbone, memory management and deployment conditions, which matter on fast streams where expensive methods may fall behind arrivals \citep{ghunaim2023real,alfarra2023evaluation}. The following studies therefore vary memory composition, memory size, label delay, context refresh rate and query batching one at a time, holding the other settings fixed.

\textbf{Memory composition and size.} Memory size has a substantially larger effect on predictive performance than the choice of memory management policy (Table~\ref{tab:budget}). Replacing either existing memory management policy with FIFO on the same backbone produces little change in predictive performance at any budget while reducing serving cost (Table~\ref{tab:metrics}). Increasing the budget improves predictive performance more substantially, but also raises latency. TabICL gains most at smaller budgets before serving cost rises sharply, while TabPFN shows a more gradual increase in cost. Larger memories are not uniformly better, as the abrupt-drift stream reaches its best performance below the largest budget. These results favour tuning memory size before introducing more elaborate memory management.

\begin{table}[!ht]
\centering
\caption{Accuracy and FIFO latency across memory budgets on a real and a synthetic abrupt-drift stream. $\Delta$ is FIFO minus the paired memory management policy. Best per stream in bold.}
\label{tab:budget}
\small
\setlength{\tabcolsep}{4.5pt}
\begin{tabular}{lrrrrrrrrr}
\toprule
& & \multicolumn{4}{c}{TabICL\,v2} & \multicolumn{4}{c}{TabPFN\,v1} \\
\cmidrule(lr){3-6}\cmidrule(lr){7-10}
Stream & $B$ & FIFO & CURE & $\Delta$ & ms/inst. & FIFO & DualFIFO & $\Delta$ & ms/inst. \\
\midrule
\multirow{4}{*}{\texttt{noaa}} & $100$ & .788 & .787 & $+.001$ & 21.9 & .784 & .785 & $-.001$ & \textbf{28.3} \\
 & $250$ & .796 & .798 & $-.002$ & \textbf{21.0} & .792 & .785 & $+.006$ & 30.8 \\
 & $500$ & .814 & .813 & $+.000$ & 24.0 & .808 & .808 & $+.000$ & 41.0 \\
 & $1{,}000$ & \textbf{.818} & \textbf{.819} & $-.001$ & 71.8 & \textbf{.814} & \textbf{.814} & $+.001$ & 99.1 \\
\midrule
\multirow{4}{*}{\texttt{agr\_a}} & $100$ & .921 & .918 & $+.003$ & \textbf{17.3} & .901 & .902 & $-.001$ & \textbf{24.9} \\
 & $250$ & .935 & .935 & $+.000$ & 18.4 & .924 & .922 & $+.001$ & 29.7 \\
 & $500$ & \textbf{.938} & \textbf{.938} & $+.000$ & 22.8 & \textbf{.924} & \textbf{.925} & $-.001$ & 41.6 \\
 & $1{,}000$ & .935 & .937 & $-.002$ & 71.4 & .919 & .920 & $+.000$ & 113.0 \\
\bottomrule
\end{tabular}
\end{table}

\textbf{Label delay.} Delayed labels reduce predictive performance across the evaluated learners, while TabFM+FIFO retains the highest mean performance at every delay (Figure~\ref{fig:delay}, with per-stream values in Table~\ref{tab:delay} of Appendix~\ref{app:results-delay}). In relative terms the in-context methods lose about as much as the strongest ensembles and more than the weakest learners, whose lower starting accuracy leaves less to lose, so their advantage under delay comes from starting higher rather than from degrading less. On noaa the TFMs at the longest delay still match the best streaming ensemble with immediate labels, although this does not hold on the other streams. CURE becomes beneficial on the smart-meter stream, where it increasingly outperforms FIFO on the same backbone as delay grows and surpasses TabFM+FIFO at the longest delay. This suggests that selecting among older labelled examples can help when recent feedback is unavailable. The benefit is confined to this stream, which also accounts for much of the overall decline, so we do not claim that delayed feedback generally favours elaborate memory management.

\begin{figure}[!ht]
    \centering
    \includegraphics[width=\linewidth]{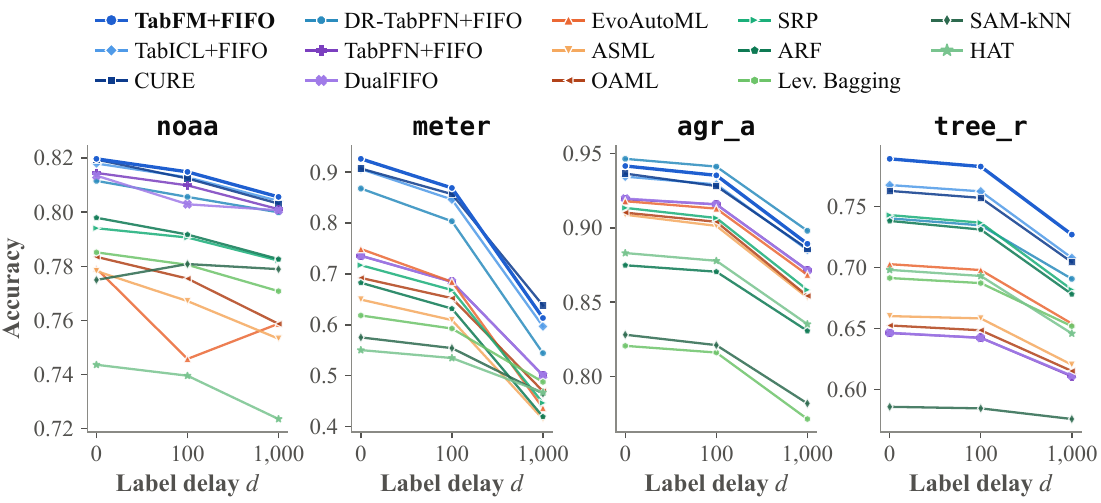}
    \caption{Accuracy of all methods as the label delay $d$ grows, where each label arrives $d$ instances after its prediction.}
    \label{fig:delay}
\end{figure}

\textbf{Context refresh rate and query batching.} Deployment levers can spread the cost of encoding the memory over more than one prediction, one across time and one across queries. Refreshing the context every $K$ instances while still predicting on every instance saves only a small share of latency at the first interval with little loss in predictive performance, and further reductions in update frequency buy almost no additional saving while degrading predictive performance, most sharply on the smart-meter stream (Table~\ref{tab:staleness}). Serving $q$ queries against a shared context instead amortises the encoding across simultaneous predictions, and per-instance serving time falls substantially on every backbone while predictive performance remains similar on the recurring-drift stream and declines more noticeably on the smart-meter stream at larger batches (Table~\ref{tab:batch}). Neither lever removes the underlying cost, since a refresh only stores the new memory while every prediction call re-encodes it.

\begin{table}[!ht]
    \centering
    \caption{TabICL with FIFO when the context is refreshed every $K$ instances but queried on every instance. Column headers give $K$, and the best per row and metric is in bold.}
    \label{tab:staleness}
    \small
    \setlength{\tabcolsep}{4.5pt}
    \begin{tabular}{lrrrrrrrrrrrr}
    \toprule
    & \multicolumn{4}{c}{Context updates} & \multicolumn{4}{c}{Accuracy} & \multicolumn{4}{c}{Time (ms/inst.)} \\
    \cmidrule(lr){2-5}\cmidrule(lr){6-9}\cmidrule(lr){10-13}
    Stream & $1$ & $10$ & $100$ & $1{,}000$ & $1$ & $10$ & $100$ & $1{,}000$ & $1$ & $10$ & $100$ & $1{,}000$ \\
    \midrule
    \texttt{noaa} & $18{,}059$ & $1{,}806$ & $181$ & $19$ & \textbf{.818} & .816 & .814 & .808 & 31.5 & 27.8 & 27.3 & \textbf{27.1} \\
    \texttt{meter} & $22{,}848$ & $2{,}285$ & $229$ & $23$ & \textbf{.906} & .903 & .875 & .710 & 70.3 & 62.0 & 61.3 & \textbf{60.7} \\
    \texttt{agr\_a} & $29{,}833$ & $2{,}984$ & $299$ & $30$ & \textbf{.935} & .934 & .932 & .917 & 32.4 & 28.5 & 28.0 & \textbf{27.9} \\
    \texttt{tree\_r} & $59{,}900$ & $5{,}990$ & $599$ & $60$ & .767 & \textbf{.768} & .764 & .737 & 35.0 & 31.0 & 30.7 & \textbf{30.6} \\
    \bottomrule
    \end{tabular}
\end{table}

\begin{table}[!ht]
    \centering
    \caption{Query batching with $q$ queries sharing one context, for each backbone with FIFO. Best per row and metric in bold.}
    \label{tab:batch}
    \footnotesize
    \setlength{\tabcolsep}{1.3pt}
    \renewcommand{\arraystretch}{0.9}
    \begin{tabular}{llrrrrrr}
    \toprule
    & & \multicolumn{3}{c}{Time (ms/inst.)} & \multicolumn{3}{c}{Accuracy} \\
    \cmidrule(lr){3-5}\cmidrule(lr){6-8}
    Backbone & Stream & $q{=}1$ & $q{=}10$ & $q{=}100$ & $q{=}1$ & $q{=}10$ & $q{=}100$ \\
    \midrule
    TabICL\,v2 & \texttt{meter}
    & 88.17 & 13.08 & \textbf{1.89}
    & \textbf{.906} & .903 & .875 \\
    & \texttt{tree\_r}
    & 110.66 & 7.69 & \textbf{0.92}
    & .767 & \textbf{.768} & .764 \\
    \midrule
    TabPFN\,v1 & \texttt{meter}
    & 371.75 & 50.47 & \textbf{5.58}
    & \textbf{.736} & .735 & .709 \\
    & \texttt{tree\_r}
    & 246.53 & 11.75 & \textbf{1.52}
    & \textbf{.646} & \textbf{.646} & .644 \\
    \midrule
    DR-TabPFN & \texttt{meter}
    & 1367.64 & 570.62 & \textbf{55.54}
    & \textbf{.868} & .866 & .835 \\
    & \texttt{tree\_r}
    & 552.59 & 125.13 & \textbf{12.46}
    & \textbf{.740} & .739 & .735 \\
    \midrule
    TabFM & \texttt{meter}
    & 658.69 & 142.97 & \textbf{16.19}
    & \textbf{.926} & .924 & .896 \\
    & \texttt{tree\_r}
    & 295.97 & 47.03 & \textbf{5.34}
    & \textbf{.789} & \textbf{.789} & .784 \\
    \bottomrule
    \end{tabular}
\end{table}

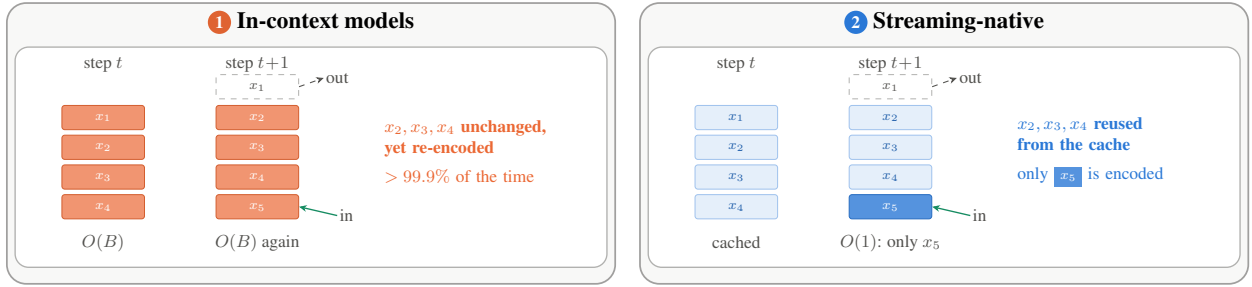
\begin{figure}[!ht]
    \centering
    \resizebox{\textwidth}{!}{% Why the cost is structural, drawn in the same language as fig_overview_2.tex:
% a soft grey panel per alternative, each holding one white content card, with a
% numbered badge in the panel title. Rows are numbered so the change between t and
% t+1 is visible: x1 is evicted, x5 arrives, x2..x4 are the SAME rows at both steps.
% The two panels then differ only in what happens to those unchanged rows.
% Needs colour macros cBlue, cAqua, cOrange, cInk, cMuted, cFaint and the tikz
% libraries positioning, fit, arrows.meta, calc, backgrounds.
\begin{tikzpicture}[
  font=\small,
  >={Stealth[length=4pt,width=3.5pt]},
  panel/.style={draw=cMuted!55, line width=.9pt, rounded corners=10pt, inner sep=0pt,
                fill=cFaint!60},
  sub/.style={draw=cMuted!45, line width=.7pt, rounded corners=7pt, inner sep=0pt, fill=white},
  ptitle/.style={font=\large\bfseries, inner sep=0pt},
  lbl/.style={font=\footnotesize, color=cMuted, align=center, inner sep=0pt},
  num/.style={circle, fill=#1, text=white, inner sep=0.5pt, minimum size=12pt,
              font=\footnotesize\bfseries},
  cell/.style={draw=cMuted!50, minimum width=15mm, minimum height=4.4mm,
               rounded corners=1.5pt, inner sep=0pt, font=\scriptsize},
  hot/.style={cell, fill=cOrange!70, draw=cOrange!90!black, text=white},
  cold/.style={cell, fill=cBlue!12, draw=cBlue!45, text=cBlue!80!black},
  new/.style={cell, fill=cBlue!80, draw=cBlue!90!black, text=white},
  gone/.style={cell, draw=cMuted!45, dashed, text=cMuted},
  evict/.style={->, draw=cMuted, line width=.55pt, dashed},
  arrive/.style={->, draw=cAqua!80!black, line width=.7pt},
  note/.style={align=left, inner sep=0pt, text width=3.9cm, font=\footnotesize},
]
\def\ytop{2.9}\def\ybot{-2.2}

% ---- containers ---------------------------------------------------------------
\begin{scope}[on background layer]
  \node[panel, fit={(0,\ytop) (11.05,\ybot)}] (P1) {};
\end{scope}
\begin{scope}[on background layer]
  \node[panel, fit={(11.5,\ytop) (22.55,\ybot)}] (P2) {};
\end{scope}
\begin{scope}[on background layer]
  \node[sub, fit={(0.15,2.1) (10.90,-1.9)}] (C1) {};
\end{scope}
\begin{scope}[on background layer]
  \node[sub, fit={(11.65,2.1) (22.40,-1.9)}] (C2) {};
\end{scope}

\node[ptitle] at (5.525,\ytop-0.35)
  {\tikz[baseline=-3pt]\node[num=cOrange!95!black]{1}; In-context models};
\node[ptitle] at (17.025,\ytop-0.35)
  {\tikz[baseline=-3pt]\node[num=cBlue]{2}; Streaming-native};

% =============================================================================
% PANEL 1 : re-encode everything
% =============================================================================
\node[lbl] at (1.75,1.78) {step $t$};
\foreach \i/\y/\n in {1/0.82/x_1, 2/0.28/x_2, 3/-0.26/x_3, 4/-0.80/x_4}
  \node[hot] (a\i) at (1.75,\y) {$\n$};
\node[lbl] at (1.75,-1.45) {$O(B)$};

\node[lbl] at (4.55,1.78) {step $t\!+\!1$};
\node[gone] (g1) at (4.55,1.38) {$x_1$};
\foreach \i/\y/\n in {1/0.82/x_2, 2/0.28/x_3, 3/-0.26/x_4, 4/-0.80/x_5}
  \node[hot] (b\i) at (4.55,\y) {$\n$};
\node[lbl] at (4.55,-1.45) {$O(B)$ again};

\draw[evict] (g1.east) -- ++(0.45,0.16) node[lbl, anchor=west, inner sep=1pt] {out};
\draw[arrive] (6.00,-0.96) node[lbl, anchor=west, inner sep=1pt] {in} -- (b4.east);

\node[note, anchor=west] at (6.85,0.20)
  {\color{cOrange}\textbf{$x_2,x_3,x_4$ unchanged,}\\[1pt]
   \color{cOrange}\textbf{yet re-encoded}\\[5pt]
   {\color{cOrange}$>99.9\%$ of the time}};

% =============================================================================
% PANEL 2 : encode once, reuse
% =============================================================================
\begin{scope}[xshift=11.5cm]
\node[lbl] at (1.75,1.78) {step $t$};
\foreach \i/\y/\n in {1/0.82/x_1, 2/0.28/x_2, 3/-0.26/x_3, 4/-0.80/x_4}
  \node[cold] (c\i) at (1.75,\y) {$\n$};
\node[lbl] at (1.75,-1.45) {cached};

\node[lbl] at (4.55,1.78) {step $t\!+\!1$};
\node[gone] (h1) at (4.55,1.38) {$x_1$};
\foreach \i/\y/\n in {1/0.82/x_2, 2/0.28/x_3, 3/-0.26/x_4}
  \node[cold] (d\i) at (4.55,\y) {$\n$};
\node[new] (d4) at (4.55,-0.80) {$x_5$};
\node[lbl] at (4.55,-1.45) {$O(1)$: only $x_5$};

\draw[evict] (h1.east) -- ++(0.45,0.16) node[lbl, anchor=west, inner sep=1pt] {out};
\draw[arrive] (6.00,-0.96) node[lbl, anchor=west, inner sep=1pt] {in} -- (d4.east);

\node[note, anchor=west] at (6.85,0.20)
  {\color{cBlue}\textbf{$x_2,x_3,x_4$ reused}\\[1pt]
   \color{cBlue}\textbf{from the cache}\\[5pt]
   {\color{cBlue}only \colorbox{cBlue!80}{\color{white}\scriptsize$x_5$} is encoded}};
\end{scope}
\end{tikzpicture}}
    \caption{Repeated context encoding versus cached context reuse.}
    \label{fig:mechanism}
\end{figure}

\textbf{Fixed and per-query cost.} To characterise the batching gains, we approximate per-call serving time by $t(q)=A+q\gamma$, where $A$ is the fixed overhead and $\gamma$ the incremental cost per query, fitted to a separate sweep on \texttt{noaa}. The fixed term dominates single-instance serving time on every backbone (Table~\ref{tab:cost}). It combines context encoding with preprocessing, data transfer and execution overhead, but separate measurements bound the non-encoding part. Dispatch does not depend on the memory budget, and preprocessing and transfer grow only linearly in it, yet the TabICL and TabPFN calls grow severalfold between the smallest and the default budget (Table~\ref{tab:budget}), so most of the default-budget call is context encoding. Storing a new memory, which includes preprocessing, also takes only a small fraction of a call. The fixed term is therefore largely amortisable, although the fit does not measure the exact encoding share or the achievable cache speed-up. Together with the refresh-rate plateau, this motivates reusing context representations across predictions (Figure~\ref{fig:mechanism}).

\begin{table}[!ht]
    \centering
    \caption{Fixed cost $A$ and per-query cost $\gamma$ fitted to $t(q)=A+q\gamma$ on a separate \texttt{noaa} sweep. The lower block is derived from the fit.}
    \label{tab:cost}
    \footnotesize
    \setlength{\tabcolsep}{6pt}
    \renewcommand{\arraystretch}{0.9}
    \begin{tabular}{lrrr}
    \toprule
    & TabICL\,v2 & TabPFN\,v1 & TabFM \\
    \midrule
    Fixed cost $A$ (ms)                       & 26.1      & 59.4      & 120 \\
    Per-query cost $\gamma$ ($\mu$s)          & 13.2      & 22.3      & 117 \\
    \midrule
    Queries per unit of fixed cost $A/\gamma$ & $1{,}980$ & $2{,}670$ & $1{,}030$ \\
    Fixed share of a single call              & $>99.9\%$ & $>99.9\%$ & $>99.9\%$ \\
    \bottomrule
    \end{tabular}
\end{table}

\begin{figure}[!ht]
    \centering
    \includegraphics[width=0.72\linewidth]{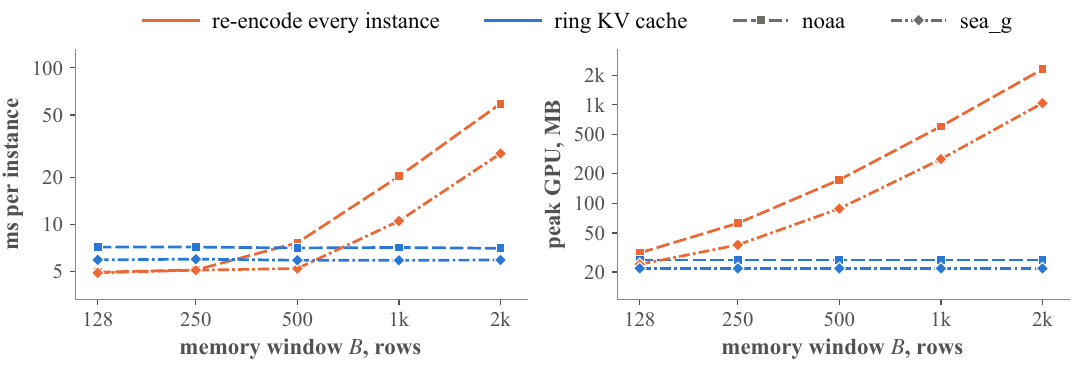}
    \caption{Per-instance serving time (left) and peak GPU memory (right) of one compact checkpoint served from a ring cache or by re-encoding every instance, as the memory window grows. The cached path stays flat, whereas re-encoding grows in both time and memory.}
    \label{fig:cache}
\end{figure}

\textbf{Cached serving.} To test whether this coupling is inherent, we pretrain compact models on a synthetic prior with drift, never on an evaluation stream, with a causal datapoint axis so that each row is encoded once on arrival and then read from a cache (Appendix~\ref{app:backbone}). Serving one checkpoint both ways on several streams, the cached path stays flat in time and GPU memory as the memory grows, while re-encoding grows superlinearly in time and close to quadratically in memory, taking up to an order of magnitude longer at the largest budget (Figure~\ref{fig:cache}). Where the checkpoint transfers, accuracy agrees at every budget except the smallest, where the gap stays small (Table~\ref{tab:cache-acc}), and on electricity both paths collapse to a constant prediction. Re-encoding remains cheaper at small memories, and the saving is far below the ceiling in Table~\ref{tab:cost}, as the models are small and the cache unoptimised. This is a proof of concept that the bottleneck is removable, not a competitive model.

\FloatBarrier
\section{Conclusions and future work}
\label{sec:discussion}

We presented a systematic study of TFMs for data streams. TFMs achieve the strongest predictive performance in our evaluation, recover faster than streaming learners after drift, and remain the most accurate under label delay, but serve more than an order of magnitude slower than the cheapest streaming ensemble. Backbone choice and memory size contribute more than the memory management policy, which paid off only under delayed labels and only on one stream. In practice, FIFO with a strong backbone and a tuned memory size is a strong default, and query batching lowers cost where latency allows. The main computational limitation is repeated context encoding at every prediction. Our compact causal models show that reusing encoded context substantially reduces serving time and memory while staying close in accuracy wherever the checkpoint transfers. However, this result is limited to the compact models we pretrain, and whether existing TFMs support similar reuse remains open. Overall, efficient streaming TFMs will require architectural changes rather than more elaborate memory management.

\textbf{Limitations.}
Our study covers stream classification only, so the findings may not carry over to regression or other tabular tasks. Our experiments are constrained by the context limits and serving costs of the evaluated TFMs, so substantially larger memories remain unexplored. In addition, context-based adaptation depends on the labelled examples available in memory, which limits adaptation under delayed feedback and changing concepts. The caching experiments are restricted to compact models and fixed attention windows, and therefore do not establish that the same efficiency gains will transfer directly to larger pretrained TFMs. Finally, unlabelled observations arriving during label delay are not used.

\textbf{Future work.}
A key direction is to develop TFMs that encode arriving examples once and reuse their representations as the stream evolves. Approaches based on local attention \citep{beltagy2020longformer}, recurrent state reuse \citep{dai2019transformerxl}, and key-value caching \citep{xiao2024streamingllm} provide useful foundations. Future models could also incorporate pretraining on drifting sequences to better balance recent information, recurring concepts, and delayed feedback. Scaling such architectures while retaining the predictive strength of current TFMs remains an important challenge.

\section*{AI Use Statement}

We used generative AI assistants, including Claude and ChatGPT, to draft and refine manuscript text, improve readability, and develop and revise code for analysis, figures, and tables. The authors reviewed all AI-assisted output and verified every reported number using the released result files and scripts. Experimental design, scientific claims, and interpretation of results remain the authors' own. We take full responsibility for the final content of this work.

%%
%% Print the bibliography
%%
\clearpage
\bibliography{references}
\bibliographystyle{arxiv}

\newpage

\appendix

\section{Streams}
\label{app:streams}

This appendix details the sources and tasks of the streams in Table~\ref{tab:streams} and the settings of the synthetic generators. The repository streams are available through the USPDS repository.\footnote{\url{https://sites.google.com/view/uspdsrepository}} The smart-meter recordings are obtained from the Dryad data repository,\footnote{\url{https://doi.org/10.5061/dryad.m0cfxpp2c}} while the person-activity recordings are available from the UCI Machine Learning Repository.\footnote{\url{https://doi.org/10.24432/C57G8X}}

\subsection{Real streams}
\label{app:streams-real}

\begin{itemize}[leftmargin=1.4em, itemsep=1pt, topsep=2pt]
\item \textbf{\texttt{electricity}:} Half-hourly records from the New South Wales electricity
market between 1996 and 1998, with eight features covering price, demand and interstate transfer.
The label is whether the price moved up or down relative to a moving average of the preceding day.
Consecutive labels are strongly autocorrelated, so repeating the last label already scores well.

\item \textbf{\texttt{noaa}:} Fifty years of daily weather measurements from one NOAA station,
with eight features such as temperature, pressure and wind speed, and a binary label for whether
rain was observed. Seasonal and climatic variation change the inputs over time, although no drift points are annotated.

\item \textbf{\texttt{meter}:} Electricity consumption recorded by smart meters in Los Alamos,
New Mexico, sampled every fifteen minutes. Each row is one household-day, so the ninety-six
features are the readings of that day in order, and the ten classes identify the household.

\item \textbf{\texttt{rialto}:} Ten buildings beside the Rialto bridge in Venice, photographed
by a fixed webcam over twenty consecutive days and encoded as a normalised twenty-seven-dimensional
colour histogram. The label is which building the image shows. Changes in illumination, weather, and time of day alter the input distribution while the building identities remain fixed.

\item \textbf{\texttt{posture\_no8}:} Position readings from four tags worn by five people
performing a scripted set of activities, giving three features, the tag's coordinates, and an
activity label. The original release has eleven activities. Following CURE~\citep{lee2026bounded}, we drop the rarest and keep ten, because TabPFN~v1 accepts at most ten classes.
\end{itemize}

\subsection{Synthetic streams}
\label{app:streams-synthetic}

All four are built through CapyMOA
\citep{gomes2024capymoa}, which wraps the MOA implementations \citep{bifet2010moa}, and none of
them appears in the pretraining prior of any model we train.

\begin{itemize}[leftmargin=1.4em, itemsep=1pt, topsep=2pt]
\item \textbf{\texttt{agr\_a}, abrupt:} The AGRAWAL generator describes loan applicants through
nine attributes such as salary, age and loan amount, and labels them by one of ten fixed
classification functions. We switch that function three times, from $1$ to $4$ to $7$ to $10$, at
instances $7{,}500$, $15{,}000$ and $22{,}500$ of a $30{,}000$-instance stream, so the concept
changes instantly at three known points while the inputs keep their distribution.

\item \textbf{\texttt{sea\_g}, gradual:} The SEA generator labels three numeric attributes by
thresholding the sum of the first two. We move from function $1$ to function $3$ over a transition
window of $10{,}000$ instances centred at the midpoint of a $100{,}000$-instance stream, so during
the transition each instance is drawn from the old or the new concept with a probability that
shifts smoothly, and there is no single point at which the concept changes.

\item \textbf{\texttt{rbf\_m}, incremental:} RandomRBF places fifty class centroids in a
ten-dimensional space and draws instances around them, with five classes. Every centroid drifts
continuously at a speed of $0.0001$ per instance, producing continuous movement of the decision boundary over $100{,}000$ instances without discrete drift points.

\item \textbf{\texttt{tree\_r}, recurring:} RandomTree labels ten attributes, five numeric and
five nominal, by a randomly generated decision tree. We build three such trees and cycle through
them with abrupt transitions every $5{,}000$ instances, four visits to each concept, giving
$60{,}000$ instances in total.
\end{itemize}

\section{Reproducibility}
\label{app:repro}

All experiments use a single-pass prequential (test-then-train)  evaluation in which predictions are made before labels are revealed. Default settings and study-specific variations are summarised in Table~\ref{tab:configs}. Streaming learners and streaming AutoML methods run on the CPU host, while TFM backbones, compact models, and backbone efficiency measurements run on the GPU hosts listed in Table~\ref{tab:platform}. Cross-method wall-clock times therefore reflect deployment cost on their respective platforms, whereas within-backbone efficiency comparisons use fixed hardware. The refresh, batching and budget sweeps are timed in separate runs whose absolute times sit below Table~\ref{tab:maincost}, so they should be compared only within each sweep. TFM backbones are isolated in separate virtual environments when required by incompatible dependencies, with TabPFN pinned to the evaluated v1 implementation. Running each paradigm on the hardware it is designed for follows the practice of published tabular benchmarks \citep{erickson2025tabarena,hollmann2023tabpfn,lourenco2025ltm}. The asymmetry favours the TFMs, so any normalisation would increase their measured cost relative to the CPU methods. Drift-Resilient TabPFN uses its released checkpoints, which share the TabPFN~v1 architecture, with four ensemble configurations. Its time covariate is the arrival index binned into equal-width domains, with each query placed one domain ahead of its memory. Its nine streams ran concurrently on the two A6000 hosts, five and four at a time, under the same harness as the other methods. TabFM is distributed without a technical report or a published parameter count, so we count parameters directly in the released checkpoints, summing tensor sizes from the safetensors header for TabFM and over the state dictionary for the other two. This gives $25.8$M parameters for TabPFN~v1, $27.6$M for TabICL~v2, and $1.64$B for the TabFM classification checkpoint, all stored in single precision. The TabFM model card additionally documents an embedding width of $256$, three column-attention blocks, three row-attention blocks, twenty-four in-context blocks, and a maximum of ten classes. Compact-model comparisons vary the datapoint-axis attention mask within each architecture family while keeping the remaining training configuration fixed. nanoTabPFN uses four layers, embedding width $160$, eight heads, and feed-forward width $320$, while nanoTabICL uses embedding width $96$, four heads, two column blocks, two row blocks, and six in-context blocks over $64$ class rows. Both are trained for $20{,}000$ steps with schedule-free AdamW on the same synthetic prior, which excludes the evaluation generators. Released result files contain the metrics, per-window histories, computational costs, memory measurements, host information, and configurations used to reproduce the reported results\footnote{\url{https://github.com/TechyNilesh/StreamTFM-Study}}.

\begin{table}[!ht]
\centering
\caption{Default evaluation settings and study-specific variations.}
\label{tab:configs}
\small
\setlength{\tabcolsep}{5pt}
\begin{tabular}{llll}
\toprule
Setting & Value & Streams & Where \\
\midrule
\multicolumn{4}{l}{\emph{Default evaluation settings}} \\
Warm-up                & 100 instances    & 9 & all \\
Evaluation window      & 1{,}000, tumbling & 9 & all \\
Label delay            & 0 (immediate)    & 9 & all \\
Retained memory        & 1{,}000 rows     & 9 & all \\
Queries per prediction & 1                & 9 & all \\
\midrule
\multicolumn{4}{l}{\emph{Varied, one quantity per study}} \\
Retained memory   & 100, 250, 500, 1{,}000        & 2 & memory size \\
Label delay       & 0, 100, 1{,}000               & 4 & label delay \\
Queries per batch & 1, 10, 100                    & 2 & query batching \\
Context refresh   & every 1, 10, 100, 1{,}000     & 4 & context refresh \\
Fine-tuning data  & 7{,}500 or 10{,}000 instances, stream prefix & 2 & \S\ref{app:finetune} \\
Attention mask    & bidirectional, causal, sliding & prior & \S\ref{app:masks} \\
Attention window  & 32, 64, 128, 256, 512         & prior & \S\ref{app:masks} \\
Serving path      & re-encode, ring cache         & 4 & \S\ref{app:cache} \\
Compact memory    & 128, 250, 500, 1{,}000, 2{,}000 & 4 & \S\ref{app:cache} \\
Recency decay, drift gate & on, off & 9 & \S\ref{app:sim} \\
\bottomrule
\end{tabular}
\end{table}
\begin{table}[!ht]
\centering
\caption{Hardware and software used for evaluation.}
\label{tab:platform}
\small
\setlength{\tabcolsep}{5pt}
\begin{tabular}{lll}
\toprule
Component & CPU host & GPU hosts \\
\midrule
Processor         & 2$\times$ AMD EPYC 7702, 128 cores & AMD EPYC 7543, 32 cores \\
Memory            & 1\,TiB DDR4                & 251\,GiB DDR4 \\
Accelerator       & none                       & NVIDIA RTX A6000, 48\,GB \\
OS                & Ubuntu 24.04.4 LTS         & Rocky Linux 9.5 \\
Kernel            & Linux 6.8.0                & Linux 5.14.0 \\
Python            & 3.14                       & 3.12 (3.11 for the v1 environment) \\
Runtime           & CapyMOA 0.13, OpenJDK 21   & CapyMOA 0.13, OpenJDK 17, PyTorch 2.13 (cu126) \\
\bottomrule
\end{tabular}
\end{table}

\section{Additional results}
\label{app:results}

\subsection{Accuracy}
\label{app:results-accuracy}

Table~\ref{tab:mainacc} gives the per-stream accuracy behind Table~\ref{tab:metrics}. TabFM+FIFO is best on every stream except \texttt{agr\_a}, where Drift-Resilient TabPFN leads, and the in-context methods separate most from the classical learners on \texttt{meter} and \texttt{rialto}.

\begin{table}[!ht]
\centering
\caption{Per-stream accuracy with mean and mean rank across methods, lower rank is better and best values are bold.}
\label{tab:mainacc}
\scriptsize
\setlength{\tabcolsep}{3pt}
\resizebox{\linewidth}{!}{%
\begin{tabular}{lccccccccc|cc}
\toprule
Method & elec & noaa & meter & rialto & posture & agr\_a & sea\_g & rbf\_m & tree\_r & Mean & Rank \\
\midrule
No-Change & .853 & .680 & .011 & .000 & .203 & .647 & .561 & .260 & .505 & .413 & 15.56 \\
Majority-Class & .575 & .686 & .011 & .100 & .333 & .632 & .673 & .343 & .523 & .431 & 15.22 \\
\midrule
HAT & .839 & .744 & .550 & .412 & .532 & .883 & .882 & .709 & .698 & .694 & 12.33 \\
ARF & .900 & .798 & .683 & .728 & .619 & .875 & .890 & .888 & .738 & .791 & 8.00 \\
SRP & .894 & .794 & .717 & .802 & .608 & .914 & .878 & .874 & .743 & .803 & 8.11 \\
Lev.\ Bagging & .890 & .785 & .618 & .612 & .600 & .821 & .886 & .861 & .691 & .752 & 10.33 \\
SAM-kNN & .825 & .775 & .575 & .817 & .594 & .828 & .865 & .930 & .586 & .755 & 11.22 \\
\midrule
ASML & .900 & .778 & .650 & .754 & .541 & .909 & .881 & .794 & .660 & .763 & 10.67 \\
EvoAutoML & .928 & .779 & .749 & .880 & .568 & .918 & .878 & .890 & .703 & .810 & 7.67 \\
OAML & .888 & .783 & .692 & .799 & .519 & .910 & .874 & .748 & .652 & .763 & 11.22 \\
\midrule
CURE & .936 & .819 & .907 & .920 & .621 & .937 & .890 & .923 & .763 & .857 & 2.89 \\
DualFIFO & .917 & .814 & .736 & .802 & .596 & .920 & .886 & .922 & .646 & .804 & 6.89 \\
\midrule
TabICL + FIFO & .944 & .818 & .906 & .917 & .629 & .935 & .891 & .932 & .767 & .860 & 2.56 \\
TabPFN + FIFO & .917 & .814 & .736 & .802 & .598 & .919 & .887 & .922 & .646 & .805 & 6.89 \\
DR-TabPFN + FIFO & .903 & .812 & .868 & .865 & .615 & \textbf{.947} & .885 & .909 & .740 & .838 & 5.33 \\
TabFM + FIFO & \textbf{.950} & \textbf{.820} & \textbf{.926} & \textbf{.924} & \textbf{.635} & .942 & \textbf{.894} & \textbf{.938} & \textbf{.789} & \textbf{.869} & \textbf{1.11} \\
\bottomrule
\end{tabular}}
\end{table}

\subsection{Label delay}
\label{app:results-delay}

Table~\ref{tab:delay} gives the per-stream accuracy behind Figure~\ref{fig:delay}. TabFM+FIFO has the best mean at every delay. Most of the decline comes from \texttt{meter}, the only stream where CURE overtakes TabFM+FIFO at the longest delay.

\begin{table}[!ht]
\centering
\caption{Accuracy under label delay $d$ on four streams, all sixteen methods. Rank is the mean per-stream rank. Best per column in bold.}
\label{tab:delay}
\small
\definecolor{tintincontext}{HTML}{E3EDFA}
\definecolor{tintAutoML}{HTML}{FCEBE5}
\definecolor{tintclassical}{HTML}{E1F5EE}
\definecolor{tinttrivial}{HTML}{EAEAE9}
\definecolor{swtabfmfifo}{HTML}{1F5FD0}
\definecolor{swtabiclv2fifo}{HTML}{4F9AE8}
\definecolor{swCURE}{HTML}{0D3B8C}
\definecolor{swdrtabpfnfifo}{HTML}{2B8CBE}
\definecolor{swtabpfnv1fifo}{HTML}{7B3FBF}
\definecolor{swLTM}{HTML}{A07BE8}
\definecolor{swevoautoml}{HTML}{EB6834}
\definecolor{swsrp}{HTML}{1BAF7A}
\definecolor{swarf}{HTML}{0E7A54}
\definecolor{swleveragingbagging}{HTML}{6EC46A}
\definecolor{swasml}{HTML}{F5A95F}
\definecolor{swoaml}{HTML}{B2431A}
\definecolor{swsamknnb1000}{HTML}{2D6A4F}
\definecolor{swhoeffdingadaptivetree}{HTML}{79C48F}
\definecolor{swmajorityclass}{HTML}{9A9A95}
\definecolor{swnochange}{HTML}{5A5A55}
\definecolor{delay100}{HTML}{EEF0F4}
\definecolor{delay1000}{HTML}{DBDEE6}
\resizebox{\linewidth}{!}{%
\begin{tblr}{colspec={lcccccccccccccccccc}, colsep=2.2pt, rowsep=0.4pt, cell{2}{2-19}={halign=c},
  column{3}={bg=delay100}, column{4}={bg=delay1000}, column{6}={bg=delay100}, column{7}={bg=delay1000}, column{9}={bg=delay100}, column{10}={bg=delay1000}, column{12}={bg=delay100}, column{13}={bg=delay1000}, column{15}={bg=delay100}, column{16}={bg=delay1000}, column{18}={bg=delay100}, column{19}={bg=delay1000}, vline{14,17}={2-Z}{0.4pt,gray}}
\toprule
& \SetCell[c=3]{c} noaa & & & \SetCell[c=3]{c} meter & & & \SetCell[c=3]{c} agr\_a & & & \SetCell[c=3]{c} tree\_r & & & \SetCell[c=3]{c} Mean & & & \SetCell[c=3]{c} Rank & & \\
\cmidrule[lr]{2-4} \cmidrule[lr]{5-7} \cmidrule[lr]{8-10} \cmidrule[lr]{11-13} \cmidrule[lr]{14-16} \cmidrule[lr]{17-19}
Method & 0 & 100 & 1000 & 0 & 100 & 1000 & 0 & 100 & 1000 & 0 & 100 & 1000 & 0 & 100 & 1000 & 0 & 100 & 1000 \\
\midrule
{\color{swnochange}\rule{0.85ex}{0.85ex}}\,No-Change & .680 & .562 & .557 & .011 & .102 & .101 & .647 & .642 & .633 & .505 & .502 & .498 & .461 & .452 & .447 & 15.8 & 15.5 & 15.8 \\
{\color{swmajorityclass}\rule{0.85ex}{0.85ex}}\,Majority-Class & .686 & .686 & .686 & .011 & .100 & .101 & .632 & .632 & .632 & .523 & .523 & .523 & .463 & .485 & .486 & 15.2 & 15.5 & 15.2 \\
\midrule
{\color{swhoeffdingadaptivetree}\rule{0.85ex}{0.85ex}}\,HAT & .744 & .740 & .724 & .550 & .535 & .467 & .883 & .878 & .835 & .698 & .693 & .646 & .719 & .711 & .668 & 11.8 & 11.8 & 10.8 \\
{\color{swarf}\rule{0.85ex}{0.85ex}}\,ARF & .798 & .792 & .783 & .683 & .632 & .419 & .875 & .871 & .831 & .738 & .731 & .678 & .773 & .756 & .678 & 8.8 & 8.8 & 9.5 \\
{\color{swsrp}\rule{0.85ex}{0.85ex}}\,SRP & .794 & .791 & .782 & .717 & .668 & .446 & .914 & .907 & .858 & .743 & .737 & .682 & .792 & .776 & .692 & 7.0 & 7.0 & 8.0 \\
{\color{swleveragingbagging}\rule{0.85ex}{0.85ex}}\,Lev.\ Bagging & .785 & .781 & .771 & .618 & .593 & .488 & .821 & .816 & .771 & .691 & .687 & .652 & .729 & .719 & .670 & 11.0 & 11.2 & 9.8 \\
{\color{swsamknnb1000}\rule{0.85ex}{0.85ex}}\,SAM-kNN & .775 & .781 & .779 & .575 & .554 & .465 & .828 & .821 & .782 & .586 & .585 & .576 & .691 & .685 & .650 & 13.2 & 12.2 & 11.5 \\
\midrule
{\color{swasml}\rule{0.85ex}{0.85ex}}\,ASML & .778 & .767 & .753 & .650 & .609 & .415 & .909 & .901 & .853 & .660 & .658 & .620 & .749 & .734 & .660 & 10.8 & 10.8 & 11.8 \\
{\color{swevoautoml}\rule{0.85ex}{0.85ex}}\,EvoAutoML & .779 & .746 & .759 & .749 & .685 & .436 & .918 & .913 & .868 & .703 & .698 & .654 & .787 & .760 & .679 & 7.5 & 8.0 & 9.2 \\
{\color{swoaml}\rule{0.85ex}{0.85ex}}\,OAML & .783 & .775 & .759 & .692 & .652 & .469 & .910 & .904 & .854 & .652 & .648 & .615 & .760 & .745 & .674 & 9.8 & 10.0 & 10.0 \\
\midrule
{\color{swCURE}\rule{0.85ex}{0.85ex}}\,CURE & .819 & .812 & .803 & .907 & .857 & \textbf{.638} & .937 & .928 & .886 & .763 & .757 & .704 & .856 & .839 & .758 & 2.5 & 3.0 & 2.5 \\
{\color{swLTM}\rule{0.85ex}{0.85ex}}\,DualFIFO & .814 & .803 & .801 & .736 & .684 & .500 & .920 & .916 & .871 & .646 & .642 & .611 & .779 & .761 & .696 & 7.0 & 7.8 & 7.2 \\
\midrule
{\color{swtabiclv2fifo}\rule{0.85ex}{0.85ex}}\,TabICL + FIFO & .818 & .813 & .804 & .906 & .846 & .597 & .935 & .929 & .885 & .767 & .762 & .708 & .857 & .838 & .748 & 3.0 & 2.5 & 2.8 \\
{\color{swtabpfnv1fifo}\rule{0.85ex}{0.85ex}}\,TabPFN + FIFO & .814 & .810 & .801 & .736 & .684 & .501 & .919 & .916 & .872 & .646 & .642 & .611 & .779 & .763 & .696 & 7.5 & 7.0 & 6.8 \\
{\color{swdrtabpfnfifo}\rule{0.85ex}{0.85ex}}\,DR-TabPFN + FIFO & .812 & .806 & .800 & .868 & .803 & .544 & \textbf{.947} & \textbf{.941} & \textbf{.898} & .740 & .734 & .691 & .842 & .821 & .733 & 4.0 & 3.8 & 3.8 \\
{\color{swtabfmfifo}\rule{0.85ex}{0.85ex}}\,TabFM + FIFO & \textbf{.820} & \textbf{.815} & \textbf{.806} & \textbf{.926} & \textbf{.869} & .613 & .942 & .935 & .889 & \textbf{.789} & \textbf{.783} & \textbf{.727} & \textbf{.869} & \textbf{.850} & \textbf{.759} & \textbf{1.2} & \textbf{1.2} & \textbf{1.5} \\
\bottomrule
\end{tblr}}

\end{table}

\subsection{Variation across streams}
\label{app:results-variation}

\begin{figure}[!ht]
    \centering
    \includegraphics[width=\linewidth]{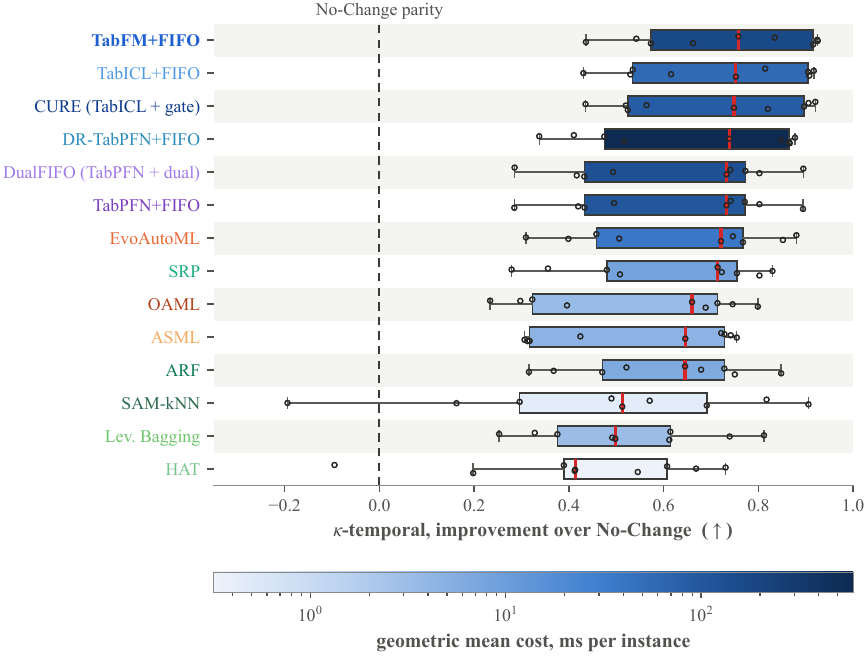}
    \caption{Temporal $\kappa_t$ across evaluation streams, with methods ordered by median. Fill colour indicates per-instance serving cost on a logarithmic scale.}
    \label{fig:box}
\end{figure}

Figure~\ref{fig:box} complements the aggregate results with the distribution of temporal $\kappa_t$ across streams. Similar medians can conceal different spreads, while differences between means and medians indicate sensitivity to individual datasets. The variation across methods, including TabFM, shows that a strong overall ranking does not imply uniform performance across stream conditions.

\subsection{Statistical comparison}
\label{app:results-stats}

\begin{figure}[!ht]
\centering
\includegraphics[width=\linewidth]{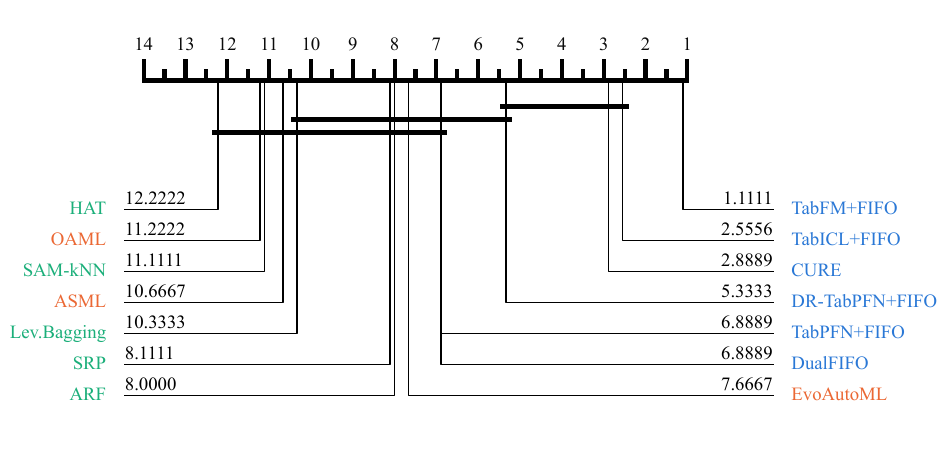}
\caption{Mean accuracy ranks with Holm-corrected two-sided Wilcoxon comparisons against the top-ranked method.}
\label{fig:cd}
\end{figure}

Figure~\ref{fig:cd} summarises accuracy ranks of the non-trivial methods, excluding No-Change and Majority-Class, so its ranks differ from those in Table~\ref{tab:metrics}. We use two-sided exact Wilcoxon signed-rank tests on paired per-stream accuracies, with Holm correction for comparisons against the top-ranked method \citep{benavoli2016should}. TabFM+FIFO outperforms every comparator on every stream except Drift-Resilient TabPFN on the abrupt-drift stream. With few streams, however, even this outcome gives a two-sided $p$-value just above the first Holm threshold, so the diagram does not separate the leading in-context methods. We therefore emphasise the descriptive result that TabFM+FIFO ranks first on all streams but one.

\FloatBarrier
\subsection{Decision boundaries on a synthetic stream}
\label{app:boundary}

\begin{figure}[!ht]
\centering
\includegraphics[width=0.9\linewidth]{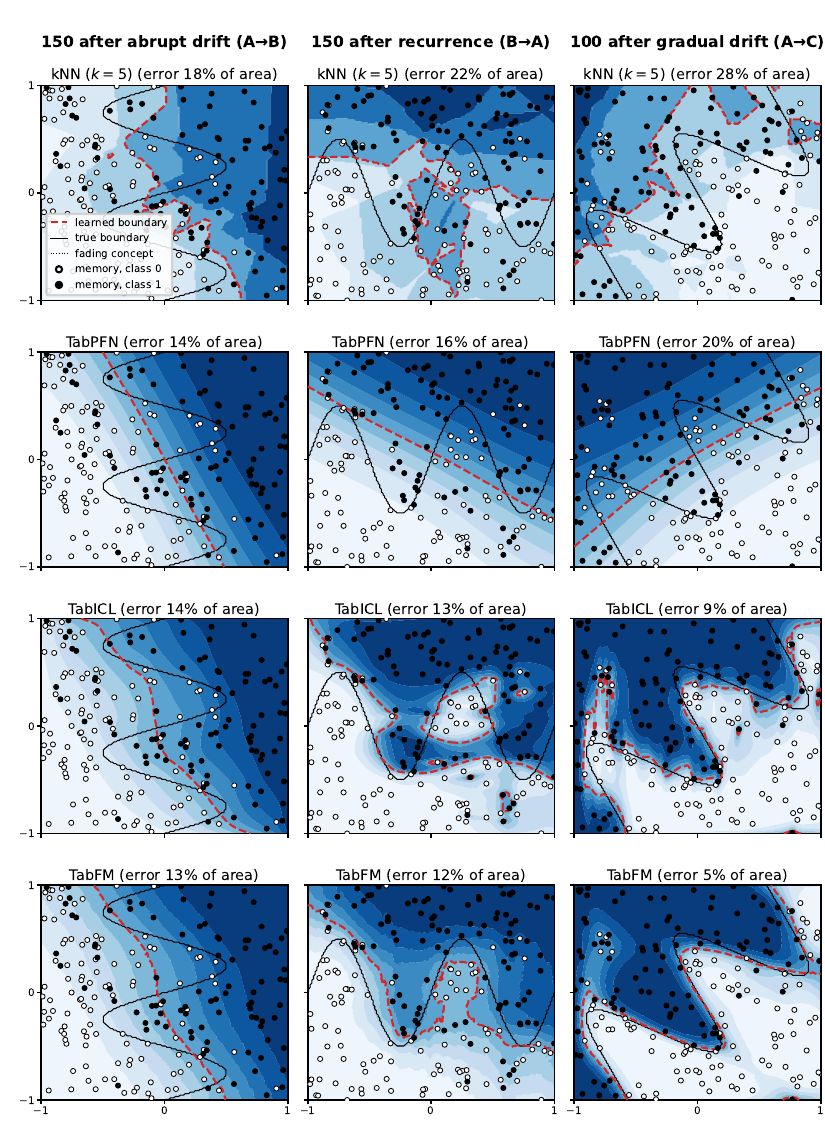}
\caption{Decision functions of kNN and the TFMs on the same FIFO memory after an abrupt drift, a recurrence and a gradual drift, with learned (red dashed) and true (black) boundaries and memory contents as dots.}
\label{fig:boundary}
\end{figure}

\begin{figure}[!ht]
\centering
\includegraphics[width=\linewidth]{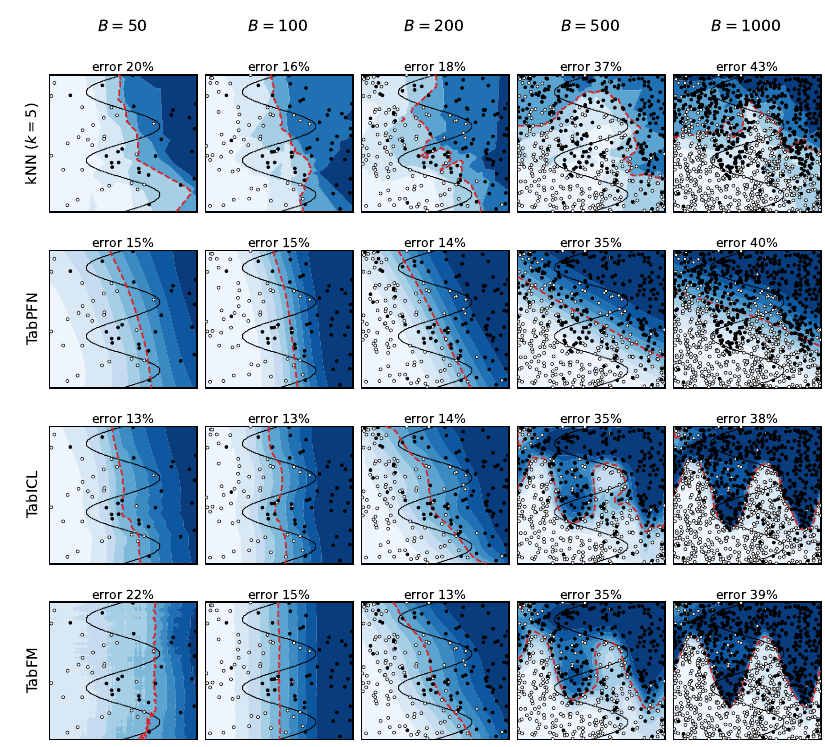}
\caption{Decision functions after the first abrupt drift for increasing FIFO memory sizes. Large memories still hold the previous concept.}
\label{fig:boundary-sizes-drift}
\end{figure}

To make the difference between in-context and instance-based learning visible, we generate a two-dimensional stream with irrelevant uniform features and label noise, whose sine-shaped concept switches abruptly to a rotation of itself, recurs, and then drifts gradually to another rotation. A kNN classifier, TabPFN~v1, TabICL~v2 and TabFM are fitted on the identical FIFO memory and evaluated on a grid over the two informative features (Figure~\ref{fig:boundary}). Because all four see the same examples, every difference comes from how they read them. The irrelevant features corrupt the distances of kNN, while the TFMs ignore them and follow the true boundary in the order of their overall accuracy. Right after each drift all four fail alike, since the memory still holds the previous concept, so under FIFO the speed of adaptation is set by memory turnover and the quality of the recovered boundary by the backbone. Memory size exposes the resulting trade-off. A larger memory sharpens the steady-state boundary of TabICL and TabFM but still holds the old concept long after a drift (Figure~\ref{fig:boundary-sizes-drift}), and recovery time grows nearly in proportion to memory size (Table~\ref{tab:boundary-recovery}). Choosing the size per window with hindsight is worth several points, and for TabICL and TabFM a selector over all sizes captures part of this headroom, whereas switching between only the smallest and largest memory does not (Table~\ref{tab:boundary-oracle}). These results come from one stream and one seed, so they indicate headroom for memory management rather than a method.

\begin{table}[!ht]
\centering
\caption{Instances after the first drift until windowed accuracy regains $90$ percent of its pre-drift level, by FIFO memory size $B$, with the slope on logarithmic axes. The smallest values sit at the measurement floor.}
\label{tab:boundary-recovery}
\small
\begin{tabular}{lrrrrrr}
\toprule
& $B{=}50$ & $100$ & $200$ & $500$ & $1{,}000$ & slope \\
\midrule
kNN ($k{=}5$) & 50 & 50 & 55 & 130 & 271 & 0.99 \\
TabPFN~v1 & 50 & 82 & 109 & 244 & 271 & 0.58 \\
TabICL~v2 & 60 & 97 & 203 & 334 & 601 & 0.67 \\
TabFM & 60 & 67 & 165 & 335 & 601 & 0.80 \\
\bottomrule
\end{tabular}
\end{table}

\begin{table}[!ht]
\centering
\caption{Mean accuracy around the first drift (all) and just after it (+300) for fixed memory sizes, oracles and selectors.}
\label{tab:boundary-oracle}
\small
\setlength{\tabcolsep}{4.5pt}
\begin{tabular}{lrrrrrrrr}
\toprule
& \multicolumn{2}{c}{kNN} & \multicolumn{2}{c}{TabPFN~v1} & \multicolumn{2}{c}{TabICL~v2} & \multicolumn{2}{c}{TabFM} \\
\cmidrule(lr){2-3}\cmidrule(lr){4-5}\cmidrule(lr){6-7}\cmidrule(lr){8-9}
& all & +300 & all & +300 & all & +300 & all & +300 \\
\midrule
Best fixed size & .766 & .760 & .822 & .780 & .878 & .787 & .886 & .780 \\
$B=50$ & .740 & .760 & .808 & .780 & .829 & .783 & .811 & .767 \\
$B=1{,}000$ & .718 & .593 & .748 & .613 & .811 & .610 & .814 & .610 \\
\midrule
Window oracle & .808 & .777 & .848 & .807 & .923 & .823 & .928 & .830 \\
One-switch oracle & .774 & .760 & .821 & .780 & .885 & .783 & .878 & .767 \\
\midrule
Selector, all sizes & .764 & .740 & .823 & .790 & .893 & .787 & .907 & .807 \\
Selector, $50$/$1{,}000$ & .758 & .753 & .816 & .780 & .868 & .763 & .873 & .770 \\
Majority vote & .779 & .713 & .817 & .757 & .888 & .773 & .888 & .757 \\
\bottomrule
\end{tabular}
\end{table}
\FloatBarrier

\subsection{Computational Cost}
\label{app:results-cost}

Table~\ref{tab:maincost} reports per-instance serving time across streams, which captures the variation in deployment cost across datasets. Table~\ref{tab:memory} reports resident memory growth per instance between the first and last evaluation windows. This measure reduces the effect of initial process allocation and model loading, although it still includes allocator and evaluation-process overhead. The reported values are process-level memory behaviour rather than isolated model state; the raw resident and GPU traces are in the released result files.

\begin{table}[!ht]
\centering
\caption{Per-stream serving time in ms per instance and mean latency rank, with lower values indicating faster serving.}
\label{tab:maincost}
\scriptsize
\setlength{\tabcolsep}{3pt}
\resizebox{\linewidth}{!}{%
\begin{tabular}{lccccccccc|cc}
\toprule
Method & elec & noaa & meter & rialto & posture & agr\_a & sea\_g & rbf\_m & tree\_r & Mean & Rank \\
\midrule
No-Change & 0.20 & 0.31 & 0.86 & 0.19 & 0.12 & 0.08 & 0.10 & 0.10 & 0.11 & 0.17 & 1.44 \\
Majority-Class & 0.30 & 0.38 & 0.67 & 0.16 & 0.22 & 0.11 & 0.09 & 0.10 & 0.12 & 0.19 & 1.78 \\
\midrule
HAT & 0.31 & 0.82 & 1.02 & 0.35 & 0.20 & 0.31 & 0.17 & 0.21 & 0.18 & 0.32 & 3.11 \\
ARF & 6.50 & 6.91 & 41.4 & 9.46 & 2.07 & 6.25 & 2.49 & 4.07 & 6.85 & 6.30 & 7.33 \\
SRP & 8.87 & 7.10 & 53.8 & 13.3 & 3.59 & 8.50 & 2.89 & 5.33 & 10.6 & 8.47 & 8.56 \\
Lev.\ Bagging & 2.44 & 2.96 & 35.2 & 6.39 & 1.64 & 2.95 & 1.00 & 2.64 & 2.14 & 3.29 & 5.78 \\
SAM-kNN & 0.31 & 0.36 & 1.96 & 0.51 & 0.38 & 0.57 & 0.27 & 0.48 & 0.26 & 0.45 & 3.67 \\
\midrule
ASML & 2.66 & 3.76 & 59.2 & 15.2 & 3.58 & 2.36 & 0.70 & 5.24 & 2.98 & 4.55 & 7.00 \\
EvoAutoML & 15.6 & 21.1 & 457 & 70.2 & 22.9 & 20.7 & 13.2 & 50.9 & 35.8 & 36.7 & 10.44 \\
OAML & 2.88 & 3.92 & 144 & 22.2 & 1.06 & 1.41 & 0.54 & 1.03 & 1.94 & 3.38 & 6.44 \\
\midrule
CURE & 31.0 & 97.9 & 150 & 116 & 65.0 & 98.8 & 93.2 & 92.0 & 101 & 87.3 & 12.22 \\
DualFIFO & 113 & 116 & 523 & 167 & 44.5 & 114 & 99.6 & 124 & 117 & 127 & 13.67 \\
\midrule
TabICL + FIFO & 30.7 & 71.8 & 88.2 & 83.2 & 62.5 & 71.4 & 28.1 & 35.2 & 111 & 58.4 & 10.89 \\
TabPFN + FIFO & 54.4 & 99.1 & 372 & 89.8 & 87.5 & 113 & 44.7 & 65.8 & 247 & 103 & 12.78 \\
DR-TabPFN + FIFO & 542 & 509 & 1368 & 628 & 471 & 540 & 486 & 614 & 553 & 599 & 16.00 \\
TabFM + FIFO & 122 & 124 & 659 & 157 & 114 & 129 & 114 & 133 & 296 & 167 & 14.89 \\
\bottomrule
\end{tabular}}
\end{table}

\begin{table}[!ht]
\centering
\caption{Per-stream resident memory growth in KB per instance and mean rank, with lower values indicating less growth. Dashes mark runs without a usable memory trace; means and ranks use the available streams.}
\label{tab:memory}
\scriptsize
\setlength{\tabcolsep}{3pt}
\resizebox{\linewidth}{!}{%
\begin{tabular}{lccccccccc|cc}
\toprule
Method & elec & noaa & meter & rialto & posture & agr\_a & sea\_g & rbf\_m & tree\_r & Mean & Rank \\
\midrule
No-Change & 0.56 & 2.65 & 1.22 & 6.30 & 1.83 & 0.07 & \textbf{0.01} & \textbf{0.01} & 0.07 & 1.41 & 4.44 \\
Majority-Class & 0.74 & 2.77 & 30.3 & 5.97 & \textbf{0.18} & 0.18 & 0.02 & 0.05 & \textbf{0.05} & 4.48 & 5.11 \\
\midrule
HAT & 4.88 & 3.79 & 33.6 & 2.16 & 0.80 & 4.27 & 1.02 & 1.28 & 5.27 & 6.34 & 9.89 \\
ARF & 0.63 & 6.87 & 4.83 & 0.56 & 4.41 & 16.6 & 7.11 & 10.3 & 11.1 & 6.93 & 11.22 \\
SRP & 11.9 & 4.88 & 16.1 & 6.35 & 5.57 & 23.4 & 7.15 & 6.46 & 14.8 & 10.7 & 14.00 \\
Lev.\ Bagging & 1.09 & 22.8 & 2.49 & \textbf{0.24} & 4.24 & 6.07 & 7.08 & 5.06 & 2.10 & 5.69 & 10.11 \\
SAM-kNN & 7.67 & 13.8 & --- & 2.82 & 1.46 & 9.72 & 2.83 & 2.93 & 9.10 & 4.86 & 10.44 \\
\midrule
ASML & \textbf{0.28} & 2.89 & 1.12 & 3.54 & 0.26 & 0.27 & 0.17 & 0.17 & --- & \textbf{0.97} & \textbf{4.22} \\
EvoAutoML & 1.65 & 6.73 & 43.3 & 3.99 & 0.48 & 2.67 & 0.64 & 1.88 & 6.90 & 7.58 & 10.00 \\
OAML & 1.70 & 3.43 & 25.7 & 3.87 & 1.77 & 0.05 & 0.13 & 0.14 & 0.20 & 4.11 & 7.22 \\
\midrule
CURE & 1.69 & 3.60 & 30.0 & 9.36 & 1.68 & --- & 1.94 & 1.96 & 3.26 & 5.94 & 9.89 \\
DualFIFO & 2.92 & 3.15 & 32.7 & 7.34 & 2.21 & --- & 0.70 & 0.72 & 0.92 & 5.63 & 9.78 \\
\midrule
TabICL + FIFO & 1.81 & 6.72 & \textbf{0.92} & 2.29 & 2.19 & 19.5 & 0.02 & 0.10 & 0.10 & 3.73 & 7.50 \\
TabPFN + FIFO & 1.32 & 3.39 & 4.43 & 6.71 & 1.96 & --- & 0.62 & 0.60 & 0.11 & 2.13 & 7.11 \\
DR-TabPFN + FIFO & 1.64 & 1.78 & 30.9 & 4.30 & 3.71 & 3.00 & 1.15 & 1.21 & 0.80 & 5.38 & 9.00 \\
TabFM + FIFO & 1.65 & \textbf{1.70} & 30.9 & 7.24 & 1.50 & \textbf{0.02} & 0.02 & 0.05 & 0.09 & 4.80 & 6.06 \\
\bottomrule
\end{tabular}}
\end{table}

\subsection{Cost at cloud prices}
\label{app:cost-price}

Table~\ref{tab:price} converts the geometric-mean serving times of Table~\ref{tab:maincost} into cloud cost per million instances. The CPU methods are charged for a full 8-vCPU on-demand instance at \$0.384 per hour, although most use a single core, and the TFMs are charged the median on-demand price of an RTX A6000 across thirteen providers, \$0.56 per hour, both as listed on 9 September 2026. Charging the CPU methods for a whole instance is conservative for our claim, since it raises their cost. The ordering of Figure~\ref{fig:pareto} is unchanged: the least expensive TFM costs about ten times the most expensive classical ensemble and over two hundred and fifty times the Hoeffding tree.

\begin{table}[!ht]
\centering
\caption{Serving cost at on-demand cloud prices, derived from Table~\ref{tab:maincost}. Prices are the AWS m6i.2xlarge rate and the median RTX A6000 rate on 9 September 2026 and are indicative only.}
\label{tab:price}
\small
\setlength{\tabcolsep}{6pt}
\begin{tabular}{lrlr}
\toprule
Method & ms/inst. & Host & USD per $10^{6}$ inst. \\
\midrule
HAT           & 0.32 & CPU & 0.03 \\
SAM-kNN       & 0.45 & CPU & 0.05 \\
Lev.\ Bagging & 3.29 & CPU & 0.35 \\
OAML          & 3.38 & CPU & 0.36 \\
ASML          & 4.55 & CPU & 0.49 \\
ARF           & 6.30 & CPU & 0.67 \\
SRP           & 8.47 & CPU & 0.90 \\
EvoAutoML     & 36.7 & CPU & 3.91 \\
\midrule
TabICL + FIFO & 58.4 & GPU & 9.08 \\
CURE          & 87.3 & GPU & 13.6 \\
TabPFN + FIFO & 103  & GPU & 16.0 \\
DualFIFO      & 127  & GPU & 19.8 \\
TabFM + FIFO  & 167  & GPU & 26.0 \\
DR-TabPFN + FIFO & 599 & GPU & 93.2 \\
\bottomrule
\end{tabular}
\end{table}

\section{Backbone studies}
\label{app:backbone}

\subsection{Fine-tuning}
\label{app:finetune}

We test whether prefix-based fine-tuning improves TabICL beyond context adaptation. Table~\ref{tab:finetune} compares the pretrained and selected fine-tuned checkpoints using FIFO memory, with tuning and scored windows kept disjoint. The tuning prefix includes a held-out validation segment, and the selected checkpoint is evaluated only on the remaining stream. Fine-tuning produces negligible changes and no clear improvement in either case, although this result is specific to the evaluated prefixes, training budget, and selection procedure.

\begin{table}[!ht]
\centering
\caption{TabICL performance before and after prefix-based fine-tuning with FIFO memory.}
\label{tab:finetune}
\small
\setlength{\tabcolsep}{4pt}
\begin{tabular}{lrrr|rrr|rrr}
\toprule
& & \multicolumn{2}{c|}{Windows} & \multicolumn{3}{c|}{Accuracy} & \multicolumn{3}{c}{$\kappa_t$} \\
\cmidrule(lr){3-4}\cmidrule(lr){5-7}\cmidrule(lr){8-10}
Stream & prefix & tuning & scored & pretrained & updated & $\Delta$ & pretrained & updated & $\Delta$ \\
\midrule
electricity & $10{,}000$ & 10 & 36 & \textbf{.9394} & .9388 & $-.0006$ & $\mathbf{+.5639}$ & $+.5600$ & $-.0039$ \\
agr\_a      & $7{,}500$  & 8  & 22 & .9404 & \textbf{.9406} & $+.0002$ & $\mathbf{+.6570}$ & $+.6566$ & $-.0004$ \\
\bottomrule
\end{tabular}
\end{table}

\subsection{Similarity instead of attention}
\label{app:sim}

To test whether the backbone advantage can be reproduced without learned attention, we replace it with a similarity-based predictor that stores labelled rows and applies cosine-similarity-weighted voting with recency decay and a drift-dependent gate. Table~\ref{tab:sim} compares this predictor using raw features and frozen TabICL row representations. Raw features perform better on most streams, while both variants remain below TabFM+FIFO. The frozen representations improve performance only on the smart-meter stream and require substantially greater serving time, suggesting that representations learned for use within the backbone do not necessarily transfer well to cosine-similarity voting. Table~\ref{tab:simablate} further examines the contribution of recency decay and the drift gate. Recency weighting improves overall performance, particularly on several real streams, while the drift gate is more beneficial on the synthetic drifting streams and shows mixed effects on real data. These results indicate that similarity-based adaptation captures some streaming behaviour but does not reproduce the predictive advantage of the full backbone.

\begin{table}[!ht]
\centering
\caption{Accuracy and serving time for similarity-based prediction using raw features, frozen TabICL representations, and TabFM+FIFO.}
\label{tab:sim}
\small
\setlength{\tabcolsep}{9pt}
\begin{tabular}{lrrr}
\toprule
Stream & raw features & TabICL row repr. & TabFM + FIFO \\
\midrule
rbf\_m       & .924 & .720 & \textbf{.938} \\
sea\_g       & .862 & .783 & \textbf{.894} \\
agr\_a       & .838 & .805 & \textbf{.942} \\
electricity  & .802 & .774 & \textbf{.950} \\
noaa         & .745 & .705 & \textbf{.820} \\
rialto       & .618 & .539 & \textbf{.924} \\
tree\_r      & .610 & .596 & \textbf{.789} \\
posture\_no8 & .532 & .509 & \textbf{.635} \\
meter        & .527 & .662 & \textbf{.926} \\
\midrule
mean         & .717 & .677 & \textbf{.869} \\
ms/inst.     & \textbf{0.24} & 15.3 & 167.4 \\
\bottomrule
\end{tabular}
\end{table}

\begin{table}[!ht]
\centering
\caption{Ablation of recency decay and the drift gate in the raw-feature similarity predictor. $\Delta$ is off minus on, so negative values favour the component.}
\label{tab:simablate}
\small
\setlength{\tabcolsep}{8pt}
\begin{tabular}{lrrr|rr}
\toprule
& \multicolumn{3}{c|}{recency decay} & \multicolumn{2}{c}{drift gate} \\
\cmidrule(lr){2-4}\cmidrule(lr){5-6}
Stream & off & on & $\Delta$ & off & $\Delta$ \\
\midrule
\multicolumn{6}{l}{\emph{Real streams}} \\
rialto       & .540 & .618 & $-.078$ & .659 & $+.042$ \\
meter        & .503 & .527 & $-.024$ & .534 & $+.006$ \\
electricity  & .789 & .802 & $-.013$ & .815 & $+.013$ \\
noaa         & .743 & .745 & $-.002$ & .744 & $-.001$ \\
posture\_no8 & .533 & .532 & $+.001$ & .503 & $-.029$ \\
\midrule
\multicolumn{6}{l}{\emph{Synthetic drifting streams}} \\
agr\_a       & .837 & .838 & $-.001$ & .810 & $-.028$ \\
sea\_g       & .863 & .862 & $+.001$ & .852 & $-.010$ \\
rbf\_m       & .924 & .924 & $+.001$ & .910 & $-.014$ \\
tree\_r      & .611 & .610 & $+.000$ & .584 & $-.027$ \\
\midrule
mean         & .705 & .717 & $-.013$ & .712 & $-.005$ \\
\bottomrule
\end{tabular}
\end{table}

\subsection{Pre-training small models}
\label{app:native}

We pre-train compact \textit{nanoTabPFN} \citep{pfefferle2025nanotabpfn}\footnote{\url{https://github.com/automl/nanoTabPFN}} and \textit{nanoTabICL} \citep{soda2026nanotabicl}\footnote{\url{https://github.com/soda-inria/nanotabicl}}
models to test whether restricted context interactions enable more efficient streaming inference. Bidirectional attention allows changes in memory to affect all retained representations, whereas causal attention restricts each row to earlier observations and can support cached reuse when the remaining architecture is compatible. Sliding attention further limits each row to a window of $W$ observations, giving $O(W)$ attention cost per step at fixed model size. Cached serving and re-encoding are not equivalent because cached rows may retain information from observations that have since been evicted, while re-encoding uses only the current memory. Within each architecture family, the attention mask is varied while the remaining training configuration is fixed, and all models are pretrained from scratch on synthetic drifting sequences that exclude the evaluation streams.

\begin{figure}[!ht]
\centering
\resizebox{0.82\textwidth}{!}{% The three datapoint-axis masks, drawn in the same language as fig_overview_2.tex:
% one soft grey panel per mask, each holding a white content card with the attention
% matrix over B retained rows. Row = the row being encoded, column = the rows it may
% read. Needs colour macros cBlue, cOrange, cInk, cMuted, cFaint and the tikz
% libraries positioning, fit, arrows.meta, calc, backgrounds.
\begin{tikzpicture}[
  font=\small,
  >={Stealth[length=4pt,width=3.5pt]},
  panel/.style={draw=cMuted!55, line width=.9pt, rounded corners=10pt, inner sep=0pt,
                fill=cFaint!60},
  sub/.style={draw=cMuted!45, line width=.7pt, rounded corners=7pt, inner sep=0pt, fill=white},
  ptitle/.style={font=\large\bfseries, inner sep=0pt},
  num/.style={circle, fill=#1, text=white, inner sep=0.5pt, minimum size=12pt,
              font=\footnotesize\bfseries},
  lbl/.style={font=\footnotesize, color=cMuted, align=center, inner sep=0pt},
  axis/.style={font=\scriptsize, color=cMuted, inner sep=0pt},
  note/.style={font=\footnotesize, color=cMuted, align=center, anchor=north, inner sep=0pt},
  on/.style={fill=cBlue!75, draw=white, line width=.3pt},
  off/.style={fill=cFaint, draw=white, line width=.3pt},
]
\def\ytop{3.4}\def\ybot{-3.4}
\def\N{8}
\def\c{0.40}
\def\W{3}

% one matrix: #1 = left edge, #2 = top edge, #3 = rule in \i (row) and \j (col), 1 = attends
\newcommand{\maskgrid}[3]{%
  \foreach \i in {1,...,\N}{
    \foreach \j in {1,...,\N}{
      \pgfmathtruncatemacro{\onval}{#3}
      \ifnum\onval=1
        \fill[on]  ({#1+(\j-1)*\c},{#2-(\i-1)*\c}) rectangle ++(\c,-\c);
      \else
        \fill[off] ({#1+(\j-1)*\c},{#2-(\i-1)*\c}) rectangle ++(\c,-\c);
      \fi } }
  \draw[cMuted!50, line width=.4pt] (#1,#2) rectangle ++({\N*\c},{-\N*\c});
  \node[axis, anchor=south] at ({#1+\N*\c/2},{#2+0.14}) {reads row $j \rightarrow$};
  \node[axis, anchor=south, rotate=90] at ({#1-0.16},{#2-\N*\c/2}) {row $i$ being encoded};
}

% ---- containers ---------------------------------------------------------------
\begin{scope}[on background layer]
  \node[panel, fit={(0,\ytop) (4.7,\ybot)}] (P1) {};
  \node[panel, fit={(5.15,\ytop) (9.85,\ybot)}] (P2) {};
  \node[panel, fit={(10.3,\ytop) (15.0,\ybot)}] (P3) {};
\end{scope}
\begin{scope}[on background layer]
  \node[sub, fit={(0.15,2.6) (4.55,-3.1)}] (C1) {};
  \node[sub, fit={(5.30,2.6) (9.70,-3.1)}] (C2) {};
  \node[sub, fit={(10.45,2.6) (14.85,-3.1)}] (C3) {};
\end{scope}

\node[ptitle] at (2.35,\ytop-0.35)
  {\tikz[baseline=-3pt]\node[num=cOrange!95!black]{1}; Bidirectional};
\node[ptitle] at (7.50,\ytop-0.35)
  {\tikz[baseline=-3pt]\node[num=cBlue]{2}; Causal};
\node[ptitle] at (12.65,\ytop-0.35)
  {\tikz[baseline=-3pt]\node[num=cBlue]{3}; Sliding, window $W$};

% ---- 1. bidirectional ---------------------------------------------------------
\maskgrid{0.95}{1.75}{1}
\node[note] at (2.35,-1.80)
  {every row reads every other\\ evict one $\Rightarrow$ all $B$ change\\
   \textcolor{cOrange!90!black}{nothing reusable}};

% ---- 2. causal ----------------------------------------------------------------
\maskgrid{6.10}{1.75}{\j<=\i ? 1 : 0}
\node[note] at (7.50,-1.80)
  {a row reads only the past\\ arrival encodes one row\\
   \textcolor{cBlue!80!black}{earlier rows unchanged}};

% ---- 3. sliding window --------------------------------------------------------
\maskgrid{11.25}{1.75}{(\j<=\i && \j>\i-\W) ? 1 : 0}
% brace marking W under the last row
\draw[cInk, line width=.5pt]
  ({11.25+(\N-\W)*\c},{1.75-\N*\c-0.07}) -- ++(0,-0.10) -- ++({\W*\c},0) -- ++(0,0.10);
\node[axis, anchor=west] at ({11.25+\N*\c+0.08},{1.75-\N*\c-0.14}) {$W$};
\node[note] at (12.65,-1.80)
  {reads the last $W$ rows only\\ eviction beyond $W$ is free\\
   \textcolor{cBlue!80!black}{ring cache, $O(1)$ per step}};
\end{tikzpicture}}
\caption{Datapoint-axis attention masks for bidirectional, causal, and sliding attention.}
\label{fig:masks}
\end{figure}
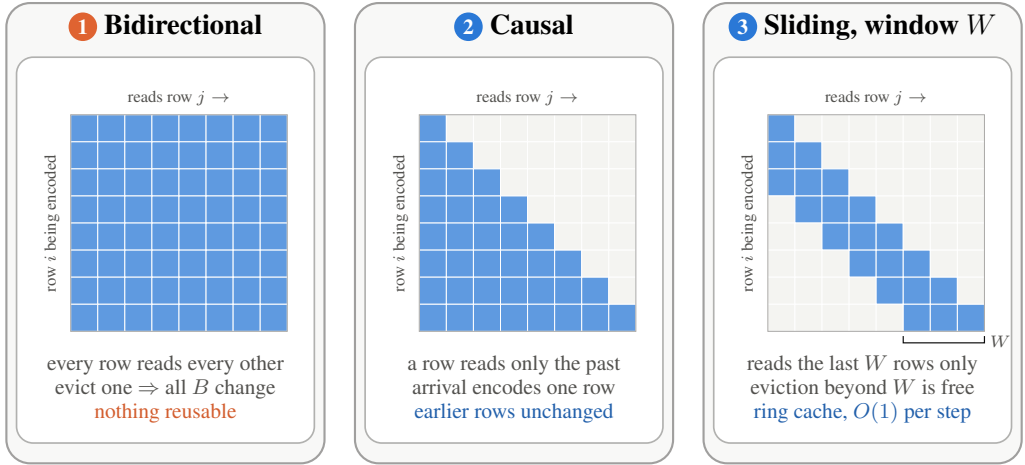

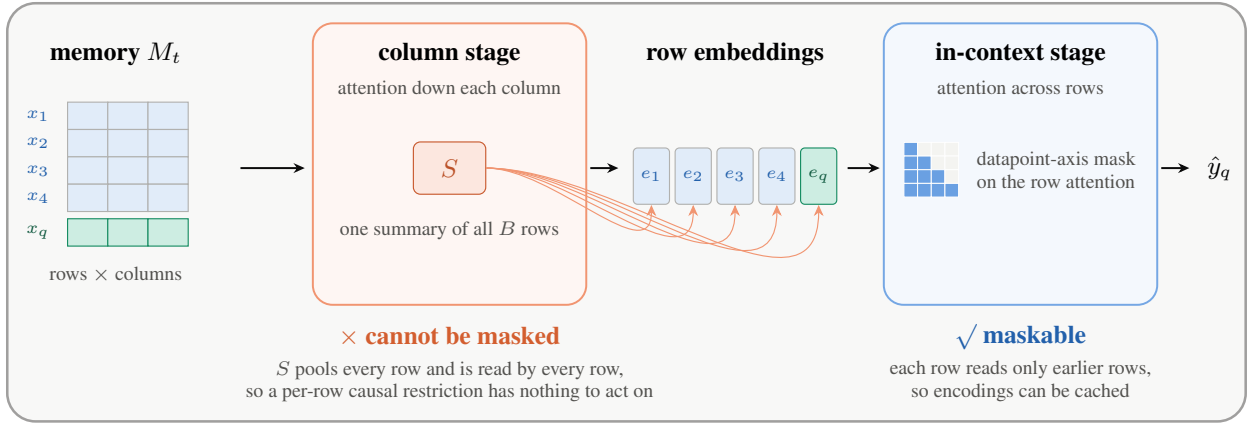
\begin{figure}[!ht]
\centering
\resizebox{\textwidth}{!}{% Why a TabICL-shaped model cannot be made cacheable by masking alone, drawn in the
% same language as fig_overview_2.tex: one soft grey panel holding a left-to-right row
% of white content cards, the two attention stages tinted by their verdict. Left to
% right: the retained rows, the column stage that pools them into a shared summary, the
% row embeddings, the in-context stage where the datapoint-axis mask lives, the
% prediction. Needs colour macros cBlue, cAqua, cOrange, cInk, cMuted, cFaint and the
% tikz libraries positioning, fit, arrows.meta, calc, backgrounds.
\begin{tikzpicture}[
  font=\small,
  >={Stealth[length=4pt,width=3.5pt]},
  panel/.style={draw=cMuted!55, line width=.9pt, rounded corners=10pt, inner sep=0pt,
                fill=cFaint!60},
  sub/.style={draw=cMuted!45, line width=.7pt, rounded corners=7pt, inner sep=0pt, fill=white},
  subo/.style={sub, draw=cOrange!70, fill=cOrange!6},
  subb/.style={sub, draw=cBlue!65, fill=cBlue!5},
  stitle/.style={font=\small\bfseries, inner sep=0pt},
  lbl/.style={font=\scriptsize, color=cMuted, align=center, inner sep=0pt},
  cell/.style={draw=cMuted!45, minimum width=5mm, minimum height=3.4mm, inner sep=0pt, font=\tiny},
  row/.style={cell, fill=cBlue!14, text=cBlue!80!black},
  qry/.style={cell, fill=cAqua!22, draw=cAqua!80!black, text=cAqua!50!black},
  emb/.style={draw=cMuted!45, rounded corners=1.5pt, minimum width=4.5mm, minimum height=7mm,
              inner sep=0pt, font=\tiny, fill=cBlue!14, text=cBlue!80!black},
  qemb/.style={emb, fill=cAqua!22, draw=cAqua!80!black, text=cAqua!50!black},
  sum/.style={draw=cOrange!85!black, fill=cOrange!16, rounded corners=2.5pt, minimum width=9mm,
              minimum height=6.5mm, font=\small\bfseries, text=cOrange!80!black},
  arr/.style={->, draw=cInk, line width=.7pt},
  fan/.style={->, draw=cOrange!70, line width=.4pt},
  bad/.style={font=\small\bfseries, color=cOrange!90!black, align=center, inner sep=0pt},
  good/.style={font=\small\bfseries, color=cBlue!80!black, align=center, inner sep=0pt},
  on/.style={fill=cBlue!70, draw=white, line width=.2pt},
  off/.style={fill=cFaint, draw=white, line width=.2pt},
]
\def\ytop{2.35}\def\ybot{-2.85}

% ---- containers ---------------------------------------------------------------
\begin{scope}[on background layer]
  \node[panel, fit={(0,\ytop) (15.4,\ybot)}] (P) {};
\end{scope}
\begin{scope}[on background layer]
  \node[subo, fit={(3.80,2.1) (7.20,-1.4)}] (C2) {};
  \node[subb, fit={(10.90,2.1) (14.30,-1.4)}] (C4) {};
\end{scope}
% ---- retained rows ------------------------------------------------------------
\node[stitle] at (1.35,1.72) {memory $M_t$};
\foreach \i/\y in {1/0.95, 2/0.61, 3/0.27, 4/-0.07}{
  \foreach \j in {1,2,3} \node[row] at ({1.00+(\j-1)*0.5},\y) {};
  \node[font=\tiny, color=cBlue!80!black, anchor=east] at (0.65,\y) {$x_\i$};
}
\foreach \j in {1,2,3} \node[qry] at ({1.00+(\j-1)*0.5},-0.50) {};
\node[font=\tiny, color=cAqua!50!black, anchor=east] at (0.65,-0.50) {$x_q$};
\node[lbl] at (1.35,-1.02) {rows $\times$ columns};

% ---- column stage -------------------------------------------------------------
\node[stitle] at (5.50,1.72) {column stage};
\node[lbl] at (5.50,1.30) {attention down each column};
\node[sum] (S) at (5.50,0.30) {$S$};
\node[lbl] at (5.50,-0.48) {one summary of all $B$ rows};

% ---- row embeddings -----------------------------------------------------------
\node[stitle] at (9.05,1.72) {row embeddings};
\foreach \i in {1,...,4} \node[emb] (e\i) at ({8.01+(\i-1)*0.52},0.20) {$e_\i$};
\node[qemb] (eq) at (10.09,0.20) {$e_q$};
% every embedding reads the one shared summary
\foreach \i in {1,...,4} \draw[fan] (S.east) to[out=0, in=-90] (e\i.south);
\draw[fan] (S.east) to[out=0, in=-90] (eq.south);

% ---- in-context stage ---------------------------------------------------------
\node[stitle] at (12.60,1.72) {in-context stage};
\node[lbl] at (12.60,1.30) {attention across rows};
\begin{scope}[shift={(11.15,0.62)}]
  \foreach \i in {1,...,4}{ \foreach \j in {1,...,4}{
    \ifnum\j>\i \fill[off] ({(\j-1)*0.17},{-(\i-1)*0.17}) rectangle ++(0.17,-0.17);
    \else       \fill[on]  ({(\j-1)*0.17},{-(\i-1)*0.17}) rectangle ++(0.17,-0.17); \fi } }
\end{scope}
\node[lbl, align=left, anchor=west] at (12.05,0.28)
  {datapoint-axis mask\\ on the row attention};

% ---- prediction ---------------------------------------------------------------
\node[font=\small, anchor=west] at (14.80,0.30) {$\hat{y}_q$};

% ---- flow ---------------------------------------------------------------------
\draw[arr] (2.90,0.30) -- (3.70,0.30);
\draw[arr] (7.25,0.30) -- (7.60,0.30);
\draw[arr] (10.45,0.30) -- (10.85,0.30);
\draw[arr] (14.35,0.30) -- (14.70,0.30);

% ---- verdicts -----------------------------------------------------------------
\node[bad] at (5.50,-1.80) {$\times$ cannot be masked};
\node[lbl] at (5.50,-2.35)
  {$S$ pools every row and is read by every row,\\ so a per-row causal restriction has nothing to act on};
\node[good] at (12.60,-1.80) {$\surd$ maskable};
\node[lbl] at (12.60,-2.35)
  {each row reads only earlier rows,\\ so encodings can be cached};
\end{tikzpicture}}
\caption{Cache dependencies in the evaluated TabICL-style architecture.}
\label{fig:tabicl}
\end{figure}

\subsection{Datapoint-axis masks}
\label{app:masks}

\begin{figure}[t]
\centering
\includegraphics[width=0.85\textwidth]{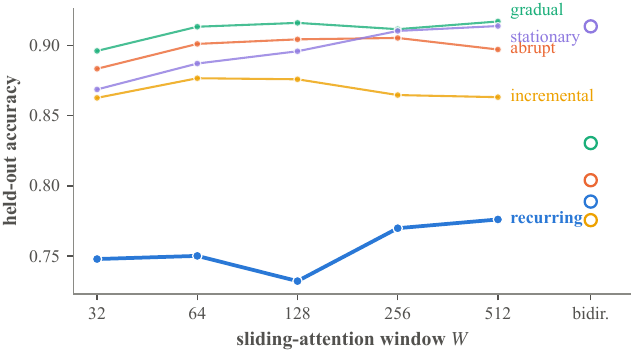}
\caption{Accuracy sensitivity to the sliding-attention window size.}
\label{fig:wsweep}
\end{figure}

\begin{figure}[t]
\centering
\includegraphics[width=\textwidth]{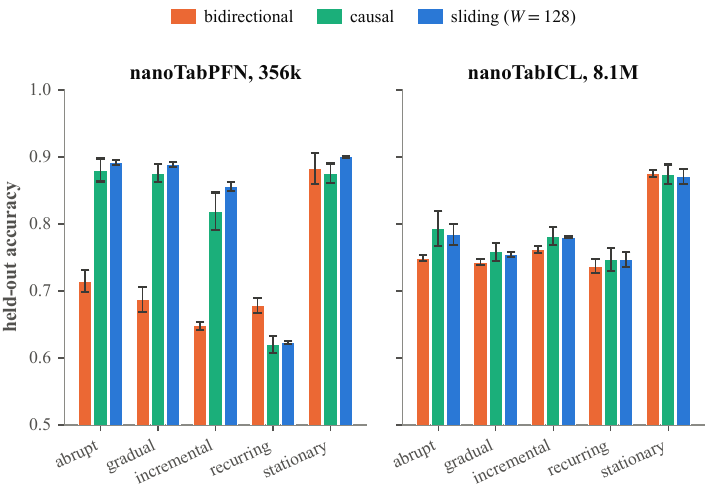}
\caption{Accuracy across bidirectional, causal, and sliding attention masks.}
\label{fig:mask}
\end{figure}

Table~\ref{tab:mask} and Figure~\ref{fig:mask} report the predictive effects of the attention masks illustrated in Figure~\ref{fig:masks}. Causal and sliding attention improve accuracy under abrupt, gradual, and incremental drift in both architecture families, with larger gains for nanoTabPFN. Recurring drift differs across architectures, while Figure~\ref{fig:tabicl} shows the shared column-stage dependencies in nanoTabICL that limit full cache compatibility. Table~\ref{tab:wsweep} and Figure~\ref{fig:wsweep} examine sensitivity to the sliding-attention window. Wider windows mainly benefit recurring and stationary sequences, while the other drift types change less beyond moderate window sizes. The mask comparison and window sweep are separate training experiments, so values should be compared within each experiment.

\begin{table}[!ht]
\centering
\caption{Accuracy across datapoint-axis attention masks and drift conditions.}
\label{tab:mask}
\small
\setlength{\tabcolsep}{7pt}
\begin{tabular}{lrrr|rrr}
\toprule
& \multicolumn{3}{c|}{nanoTabPFN, $356$k} & \multicolumn{3}{c}{nanoTabICL, $8.1$M} \\
\cmidrule(lr){2-4}\cmidrule(lr){5-7}
Drift type & bidirectional & causal & sliding & bidirectional & causal & sliding \\
\midrule
abrupt      & .715 & .880 & \textbf{.891} & .749 & \textbf{.793} & .784 \\
gradual     & .687 & .876 & \textbf{.889} & .743 & \textbf{.758} & .755 \\
incremental & .648 & .819 & \textbf{.856} & .761 & \textbf{.781} & .780 \\
recurring   & \textbf{.679} & .620 & .623 & .737 & \textbf{.747} & .747 \\
stationary  & .883 & .875 & \textbf{.900} & \textbf{.875} & .874 & .870 \\
\midrule
mean        & .722 & .814 & \textbf{.832} & .773 & \textbf{.791} & .787 \\
\bottomrule
\end{tabular}
\end{table}

\begin{table}[!ht]
\centering
\caption{Accuracy across sliding-attention window sizes for nanoTabPFN, trained separately from Table~\ref{tab:mask}.}
\label{tab:wsweep}
\small
\setlength{\tabcolsep}{7pt}
\begin{tabular}{lrrrrr|r}
\toprule
$W$ & abrupt & gradual & incremental & recurring & stationary & mean \\
\midrule
$32$  & .883 & .896 & .863 & .748 & .869 & .852 \\
$64$  & .901 & .913 & \textbf{.877} & .750 & .887 & .866 \\
$128$ & .904 & .916 & .876 & .732 & .896 & .865 \\
$256$ & \textbf{.906} & .912 & .865 & .770 & .910 & .872 \\
$512$ & .897 & \textbf{.917} & .863 & .776 & \textbf{.914} & \textbf{.874} \\
\midrule
bidirectional & .804 & .831 & .776 & \textbf{.789} & \textbf{.914} & .823 \\
\bottomrule
\end{tabular}
\end{table}

\subsection{Transfer to evaluation streams}
\label{app:transfer}

Table~\ref{tab:transfer} evaluates the compact pretrained models on evaluation streams outside the pretraining prior. Both variants outperform No-Change on several streams but perform poorly on electricity, where short-lag label autocorrelation is strong, indicating limited use of recent temporal information. A learned recency bias improves NOAA but reduces performance elsewhere, and its slopes remain at their initial values, providing no evidence that useful recency weighting was learned. Both compact models also remain below the released TabICL backbone, leaving the effects of greater capacity and broader pretraining unresolved.

\begin{table}[!ht]
\centering
\caption{Transfer accuracy and temporal $\kappa_t$ on evaluation streams. Bold marks the better compact model.}
\label{tab:transfer}
\small
\setlength{\tabcolsep}{5pt}
\begin{tabular}{lrrrrrrrr}
\toprule
& \multicolumn{2}{c}{electricity} & \multicolumn{2}{c}{noaa} & \multicolumn{2}{c}{sea\_g} & \multicolumn{2}{c}{tree\_r} \\
\cmidrule(lr){2-3}\cmidrule(lr){4-5}\cmidrule(lr){6-7}\cmidrule(lr){8-9}
Model & acc. & $\kappa_t$ & acc. & $\kappa_t$ & acc. & $\kappa_t$ & acc. & $\kappa_t$ \\
\midrule
No-Change              & .853 & \phantom{$-$}.000 & .680 & \phantom{$-$}.000 & .561 & \phantom{$-$}.000 & .505 & \phantom{$-$}.000 \\
TabICL + FIFO          & .944 & $+.616$ & .818 & $+.431$ & .891 & $+.753$ & .767 & $+.530$ \\
\midrule
sliding                & .602 & $-1.71$ & .732 & $+.160$ & \textbf{.814} & $\mathbf{+.575}$ & \textbf{.559} & $\mathbf{+.108}$ \\
sliding, learned recency bias & .428 & $-2.90$ & \textbf{.742} & $\mathbf{+.194}$ & .732 & $+.389$ & .551 & $+.091$ \\
\bottomrule
\end{tabular}
\end{table}

\subsection{Cached serving}
\label{app:cache}

Figure~\ref{fig:cache} compares cached serving with re-encoding using the same checkpoint and stream prefix. At a fixed attention window, cached serving remains nearly constant across memory budgets in both time and peak GPU memory, while re-encoding becomes increasingly expensive in both, with memory growing faster than time because the attention footprint scales with the square of the retained memory. %Re-encoding is still cheaper at small budgets because cache maintenance adds fixed overhead. Predictive performance is similar where caching is beneficial, although the two serving paths are not mathematically equivalent because cached representations may preserve information from previously evicted rows. The results demonstrate improved scaling for this compact model but do not establish the speed-up attainable by released TFMs. Table~\ref{tab:cache-acc} reports both paths at every budget on four streams. Cached accuracy is constant across budgets because the sliding window of $128$ rows fixes the effective history, whereas re-encoding conditions on the full retained memory. On electricity the checkpoint does not transfer (Table~\ref{tab:transfer}) and both paths degenerate to a constant prediction, the cached path to the majority class and the re-encoded path at most budgets to the minority class, so the accuracy gap there reflects which class the model locks onto rather than the serving path. The cost curves, which do not depend on the predictions, hold on all four streams.

\begin{table}[!ht]
\centering
\caption{The same sliding-window checkpoint served from cache and by re-encoding on a stream prefix, per memory budget. $\Delta$ is cached minus re-encoded accuracy, time is per instance, and peak GPU memory is the allocator high-water mark, which depends only on the feature count and $B$. Electricity was timed on a second host of the same type.}
\label{tab:cache-acc}
\footnotesize
\setlength{\tabcolsep}{2.6pt}
\begin{tabular}{lrrrrrrrrrr}
\toprule
& & \multicolumn{3}{c}{Accuracy} & \multicolumn{2}{c}{$\kappa_t$} & \multicolumn{2}{c}{Time (ms/inst.)} & \multicolumn{2}{c}{Peak GPU (MB)} \\
\cmidrule(lr){3-5}\cmidrule(lr){6-7}\cmidrule(lr){8-9}\cmidrule(lr){10-11}
Stream & $B$ & Cached & Re-enc. & $\Delta$ & Cached & Re-enc. & Cached & Re-enc. & Cached & Re-enc. \\
\midrule
\texttt{electricity} & $128$ & .609 & .395 & $+.214$ & $-1.469$ & $-2.824$ & 8.2 & 5.6 & 26.5 & 31.3 \\
 & $250$ & .609 & .609 & $.000$ & $-1.469$ & $-1.469$ & 7.7 & 5.7 & 26.5 & 62.7 \\
 & $500$ & .609 & .404 & $+.205$ & $-1.469$ & $-2.768$ & 8.1 & 10.4 & 26.5 & 173 \\
 & $1{,}000$ & .609 & .405 & $+.205$ & $-1.469$ & $-2.764$ & 8.2 & 28.0 & 26.5 & 604 \\
 & $2{,}000$ & .609 & .430 & $+.179$ & $-1.469$ & $-2.605$ & 8.2 & 83.1 & 26.5 & 2,297 \\
\addlinespace[2pt]
\texttt{noaa} & $128$ & .750 & .766 & $-.016$ & .221 & .271 & 7.1 & 4.9 & 26.5 & 31.3 \\
 & $250$ & .750 & .754 & $-.004$ & .221 & .233 & 7.2 & 5.1 & 26.5 & 62.7 \\
 & $500$ & .750 & .752 & $-.002$ & .221 & .227 & 7.1 & 7.6 & 26.5 & 173 \\
 & $1{,}000$ & .750 & .750 & $.000$ & .221 & .222 & 7.1 & 20.3 & 26.5 & 604 \\
 & $2{,}000$ & .750 & .746 & $+.004$ & .221 & .208 & 7.0 & 59.0 & 26.5 & 2,297 \\
\addlinespace[2pt]
\texttt{agr\_a} & $128$ & .653 & .658 & $-.006$ & .223 & .236 & 7.2 & 4.8 & 27.4 & 32.8 \\
 & $250$ & .653 & .649 & $+.003$ & .223 & .215 & 7.2 & 5.1 & 27.4 & 67.5 \\
 & $500$ & .653 & .652 & $+.001$ & .223 & .220 & 7.3 & 8.0 & 27.4 & 190 \\
 & $1{,}000$ & .653 & .653 & $.000$ & .223 & .222 & 7.4 & 22.2 & 27.4 & 671 \\
 & $2{,}000$ & .653 & .655 & $-.002$ & .223 & .227 & 7.2 & 65.4 & 27.4 & 2,549 \\
\addlinespace[2pt]
\texttt{sea\_g} & $128$ & .824 & .829 & $-.004$ & .609 & .619 & 5.9 & 4.9 & 21.8 & 24.0 \\
 & $250$ & .824 & .829 & $-.004$ & .609 & .619 & 6.0 & 5.1 & 21.8 & 37.8 \\
 & $500$ & .824 & .831 & $-.006$ & .609 & .623 & 5.9 & 5.2 & 21.8 & 88.2 \\
 & $1{,}000$ & .824 & .831 & $-.006$ & .609 & .623 & 5.9 & 10.5 & 21.8 & 281 \\
 & $2{,}000$ & .824 & .829 & $-.004$ & .609 & .619 & 5.9 & 28.4 & 21.8 & 1,039 \\
\bottomrule
\end{tabular}
\end{table}

\clearpage
\section{Prior work}
\label{app:prior}

This appendix describes the methods evaluated in this study, with the limitation of each that is most relevant here.

\textbf{Memory management for TFMs.}
\emph{DualFIFO}~\citep{lourenco2025ltm} keeps a short-term FIFO buffer of recent rows and a long-term store, and when the store is full it drops the oldest row of the most represented class, so that the memory stays class-balanced. It is evaluated only on real-world streams, where the drift type is unknown, and its cost is reported with ten instances predicted per forward pass, so neither the effect of drift type nor single-instance serving cost is measured.
\emph{CURE}~\citep{lee2026bounded} splits the budget into a short bank of recent rows and a long bank. Rows leaving the short bank enter the long bank only if their prediction entropy was high, and the long bank evicts a row that has a close same-class neighbour. Its evaluation includes one stream with a known drift type, and its runtime is compared only with DualFIFO, so its behaviour across drift types and its cost relative to classical learners remain open.

\textbf{TFM backbones.}
\emph{TabPFN~v1}~\citep{hollmann2023tabpfn} is pretrained on synthetic tables and lets every context row attend to every other row. It is limited to small contexts and at most ten classes, and it is the backbone shared by DualFIFO and Drift-Resilient TabPFN.
\emph{Drift-Resilient TabPFN}~\citep{helli2024drtabpfn} pretrains the same architecture on a prior whose causal model changes over time and receives a time index as input. It is designed and evaluated for batch data split into time domains rather than for streams, so applying it to a stream requires choosing how arrival order maps to domains.
\emph{TabICL~v2}~\citep{qu2026tabiclv2} first embeds each row from its columns and then lets rows attend to one another, which scales to larger contexts than TabPFN~v1. Its column stage attends over all rows, so the context cannot be encoded incrementally.
\emph{TabFM}~\citep{google2026tabfm} alternates attention over columns and rows, compresses each row, and then attends over the compressed rows. It is released without a technical report and accepts at most ten classes.

\textbf{Stream learners.}
\emph{Hoeffding Adaptive Tree}~\citep{bifet2009hat} grows a decision tree from running statistics and replaces subtrees when an ADWIN detector signals change. As a single tree, it adapts slowly after a drift.
\emph{Adaptive Random Forest}~\citep{gomes2017arf} and \emph{Streaming Random Patches}~\citep{gomes2019srp} train ensembles of Hoeffding trees on resampled instances and random feature subsets, and swap in background trees when a detector fires. Both need many instances before their trees split, which gives them a slow cold start.
\emph{Leveraging Bagging}~\citep{bifet2010leveraging} increases the resampling weight and resets its worst member on detected change, so each reset rebuilds a member from scratch.
\emph{SAM-kNN}~\citep{losing2016samknn} keeps a short-term memory sized by recent error and a long-term memory of consistent older examples, and predicts by nearest neighbours. It adapts through stored examples like a TFM, but its distances degrade with irrelevant features.

\textbf{Online AutoML.}
\emph{OAML}~\citep{celik2023oaml} re-runs a pipeline search on a buffer when drift is detected, so it depends on the detector and adds search cost at every trigger.
\emph{EvoAutoML}~\citep{kulbach2022evoautoml} evolves a population of pipelines and replaces the weakest by mutation, with a cost that grows with the population.
\emph{ASML}~\citep{verma2024asml} searches around the current best pipeline, balancing exploration and exploitation. Like the other AutoML systems, its accuracy depends on search running alongside prediction.

\textbf{Evaluation under delay and compute limits.}
Prequential evaluation assumes that each label arrives before the next prediction~\citep{gama2013prequential}. Delayed-label protocols relax this assumption~\citep{delayed_labelling,souza2020evaluating}, and work on online continual learning and test-time adaptation shows that a method slower than the arrival rate must skip or delay updates~\citep{ghunaim2023real,alfarra2023evaluation,prabhu2023computationally}. Existing TFM stream methods assume immediate labels and report cost separately from accuracy, which our label-delay and cost studies address.

\textbf{Reusing encoded context.}
Language models avoid re-encoding their past through segment-level recurrence~\citep{dai2019transformerxl}, local attention~\citep{beltagy2020longformer} and key-value caching~\citep{xiao2024streamingllm}. These rely on causal attention over the sequence, which current TFM backbones lack, and our cached-serving proof of concept adds it to a compact TFM.

\clearpage
\section{Accuracy and cost of streaming TFMs}
\label{app:theory}

This appendix gives a simple account of why streaming TFMs are accurate and why they are costly. Let $L$ denote the number of layers, $h$ the embedding width, $W$ the attention window, $B$ the memory size and $d$ the label delay.

\subsection{Why TFMs are accurate on streams}
\label{app:theory-acc}

Let $\Pi$ be the pretraining prior and $p_\Pi(y\mid\mathbf{x},M)$ its posterior predictive distribution. For any predictor $f$, the prior-averaged log-loss satisfies
\begin{equation}
\mathcal{L}_\Pi(f)-\mathcal{L}_\Pi(p_\Pi)=\mathbb{E}_{\Pi}\Big[\mathrm{KL}\big(p_\Pi(\cdot\mid\mathbf{x},M)\,\big\|\,f(\cdot\mid\mathbf{x},M)\big)\Big]\ge 0,
\end{equation}
so pretraining drives $f_\theta$ towards $p_\Pi$ \citep{muller2022pfn}. This does not guarantee optimal accuracy on a given stream. It explains why a TFM can predict well from a small memory when the stream is well represented by the prior, and why it fails when it is not (Table~\ref{tab:transfer}).

With a full FIFO memory and label delay $d$, the memory at step $t$ holds rows $t-d-B,\dots,t-d-1$. After a single abrupt drift whose first new-concept instance is $\tau$, the share of old-concept rows is
\begin{equation}
\rho_t=1-\frac{\min\{B,\max\{0,\,t-\tau-d\}\}}{B},
\end{equation}
and the mean age of the rows in memory is $\bar a_t=d+(B+1)/2$. The memory is therefore fully replaced $B+d$ instances after the drift, without any parameter update. This is a replacement time rather than a recovery bound, since accuracy may recover earlier or remain incomplete after replacement. A larger memory or a longer delay increases the age of the context, which matches the slower recovery and the delay losses in Tables~\ref{tab:delay} and~\ref{tab:boundary-recovery}.

A Hoeffding tree splits a leaf once the observed gap $\widehat\Delta_n$ between the two best candidates exceeds
\begin{equation}
\epsilon_H(n,\delta)=R\sqrt{\frac{\ln(1/\delta)}{2n}},
\end{equation}
where $n$ counts instances at that leaf, $R$ is the range of the split criterion and $\delta$ the failure probability \citep{domingos2000vfdt}. Resolving a gap of $\eta$ therefore takes on the order of $R^2\ln(1/\delta)/(2\eta^2)$ instances per leaf, and this is paid again after every reset. This offers one explanation for the slower cold start and drift recovery of the tree ensembles in Figure~\ref{fig:dynamics}.

\subsection{Why TFMs are costly and how to reduce the cost}
\label{app:theory-cost}

For a row transformer of width $h$, with the feature count held fixed, re-encoding the memory at every prediction costs
\begin{equation}
C_{\text{re-encode}}=O\big(L(B^2h+Bh^2)\big)
\end{equation}
per instance, although only one row changes per step. The fitted serving time of Table~\ref{tab:cost} gives a per-instance cost of
\begin{equation}
\bar t(B,q)\approx\frac{A(B)}{q}+\gamma(B)
\end{equation}
when $q$ queries share one call. Within the tested range, batching approaches the fitted floor $\gamma(B)$ but does not remove $A(B)$ from each call. With causal attention over a fixed window of $W$ rows, each row is encoded once and read from a cache, so
\begin{equation}
C_{\text{cache}}=O\big(L(Wh+h^2)\big),\qquad S_{\text{cache}}=O(LWh),
\end{equation}
which no longer depends on $B$. Causality alone is not enough, since attention over the full cached history still grows with $B$. Cached and re-encoded predictions also need not agree exactly, because cached rows may reflect rows that have since been evicted (Table~\ref{tab:cache-acc}). This matches the flat cached curves in Figure~\ref{fig:cache}. Current TFMs attend over rows in both directions, so they cannot reuse encoded rows in this way without architectural change.

\end{document}